%% file: main.tex
\documentclass[10pt,twocolumn,letterpaper]{article}

\usepackage[pagenumbers]{cvpr} 

\usepackage{subcaption}
\usepackage{graphicx}
\usepackage{amsmath}
\usepackage{amssymb}
\usepackage{booktabs}
\usepackage{enumitem}

\usepackage[table]{xcolor}
\usepackage[pagebackref, breaklinks=true, colorlinks, citecolor=citecolor, linkcolor=linkcolor, bookmarks=false]{hyperref}
\definecolor{citecolor}{HTML}{0071BC}
\definecolor{linkcolor}{HTML}{ED1C24}

\usepackage[capitalize]{cleveref}
\crefname{section}{Sec.}{Secs.}
\Crefname{section}{Section}{Sections}
\crefname{table}{Tab.}{Tabs.}
\Crefname{table}{Table}{Tables}

\usepackage{algorithm}
\usepackage{bm}
\usepackage{algorithmic}
\usepackage{listings}
\usepackage{makecell}  
\usepackage{diagbox} 
\usepackage{mathtools} 
\usepackage{graphicx} 
\usepackage{multirow}

\definecolor{mygreen}{RGB}{83,161,81}
\definecolor{mypurple}{RGB}{103,81,165}
\definecolor{borderblue}{RGB}{71,117,194}
\definecolor{borderyellow}{RGB}{253,190,38}
\definecolor{detcolor}{gray}{.9}

\newcommand{\greenc}[1]{{\bf \textcolor{mygreen}{#1}}}

\definecolor{bestcolor}{gray}{.9}

\newcommand{\tablestyle}[2]{\setlength{\tabcolsep}{#1}\renewcommand{\arraystretch}{#2}\centering\footnotesize}
\newlength\savewidth\newcommand\shline{\noalign{\global\savewidth\arrayrulewidth
  \global\arrayrulewidth 1pt}\hline\noalign{\global\arrayrulewidth\savewidth}}
  
\newcolumntype{x}[1]{>{\centering\arraybackslash}p{#1pt}}
\newcolumntype{y}[1]{>{\raggedright\arraybackslash}p{#1pt}}
\newcolumntype{z}[1]{>{\raggedleft\arraybackslash}p{#1pt}}
\renewcommand{\paragraph}[1]{\vspace{1.25mm}\noindent\textbf{#1}}
\definecolor{deemph}{gray}{0.6}
\newcommand{\gc}[1]{\textcolor{deemph}{#1}}

\def\cvprPaperID{0000} 
\def\confName{ICCV}
\def\confYear{2023}

\begin{document}

\title{Going Denser with Open-Vocabulary Part Segmentation}
\author{Peize Sun$^1$~~ Shoufa Chen$^1$~~ Chenchen Zhu$^2$~~ Fanyi Xiao$^2$~~ Ping Luo$^1$~~ Saining Xie$^3$~~ Zhicheng Yan$^2$\\[0.2cm]
$^1$The University of Hong Kong \quad $^2$Meta AI \quad
$^3$New York University\\[0.2cm]
\href{https://github.com/facebookresearch/VLPart}{\small{https://github.com/facebookresearch/VLPart}}
}

\twocolumn[{
\maketitle
\vspace{-12mm}
\begin{figure}[H]
\hsize=\textwidth
\centering
\includegraphics[width=1.00\textwidth]{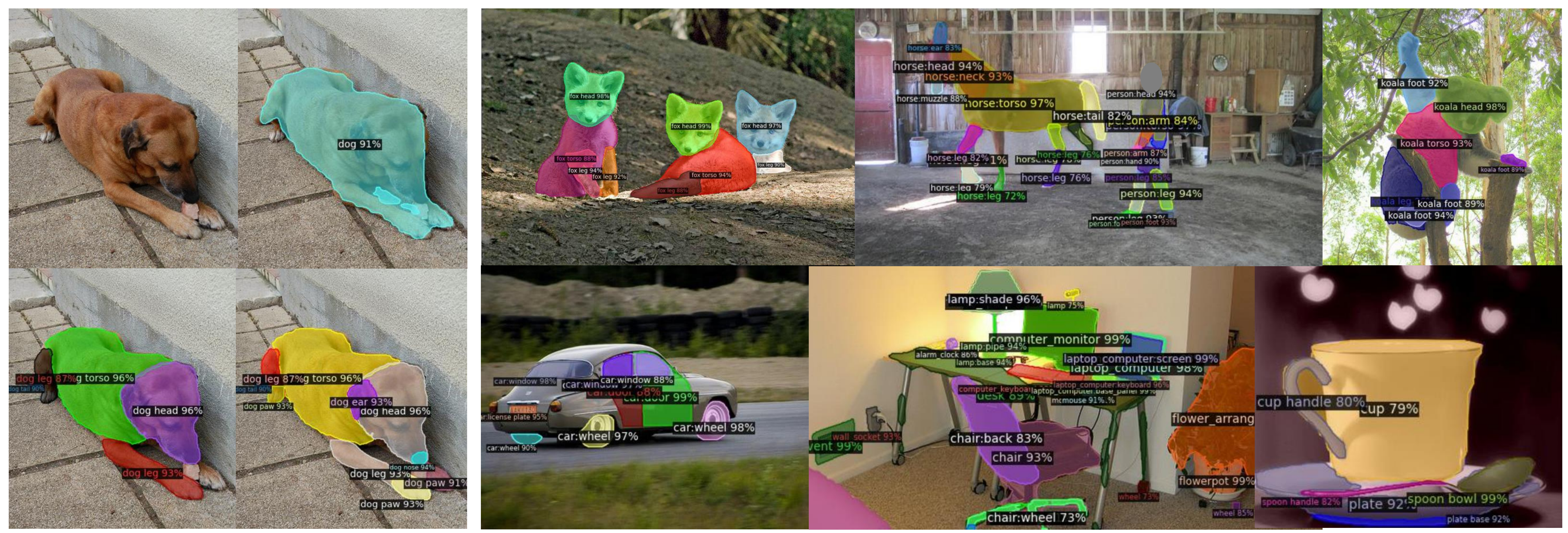} 
\vspace{-6mm}
\caption{\textbf{Examples of open-vocabulary part segmentation.} Beyond open-vocabulary object detection, we propose that the detector should be able to predict both objects and their parts. This open-world fine-grained recognition ability is in demand for an intelligent vision system but is only realized in a limited number of categories~\cite{pascalpart,partimagenet,paco} up to now. In this paper, we move forward to going denser with open-vocabulary part segmentation: Left figure shows segmenting \texttt{dog} and its parts in different granularities. Right figure demonstrates more visualization results. 
}
\label{fig:teaser}
\end{figure}
}]

\begin{abstract}
\vspace{-3mm}
Object detection has been expanded from a limited number of categories to open vocabulary. Moving forward, a complete intelligent vision system requires understanding more fine-grained object descriptions, object parts. In this paper, we propose a detector with the ability to predict both open-vocabulary objects and their part segmentation. This ability comes from two designs. First, we train the detector on the joint of part-level, object-level and image-level data to build the multi-granularity alignment between language and image. Second, we parse the novel object into its parts by its dense semantic correspondence with the base object. These two designs enable the detector to largely benefit from various data sources and foundation models. In open-vocabulary part segmentation experiments, our method outperforms the baseline by 3.3$\sim$7.3 mAP in cross-dataset generalization on PartImageNet, and improves the baseline by 7.3 novel AP$_{50}$ in cross-category generalization on Pascal Part. Finally, we train a detector that generalizes to a wide range of part segmentation datasets while achieving better performance than dataset-specific training.

\end{abstract}

\section{Introduction}
\label{intro}

Recent advances in open-vocabulary object detection~\cite{OVRCNN,regionclip,detic,glip,owlvit,vldet,fvlm,alignbag} have made surprising development in enlarging the number of object categories from a pre-determined set by training datasets~\cite{voc,coco,lvis,objects365} to any object in the open world. This is a crucial step for the vision system to make effect in the real world. Towards the next step, for a deeper understanding to object structure, mobility, functionality, and practical applications such as behavior analysis~\cite{reddy2018carfusion,yang2019parsing,ng2022animal}, robotics manipulation~\cite{rt1,nair2022r3m,partassembly,ge2022self}, image-editing~\cite{li2021partgan, rombach2022high}, only object-level perception is not sufficient, while the fine-grained recognition ability of part segmentation~\cite{pascalpart,partimagenet,paco,shapenet} is necessary.

Since a part is the fine-grained version of an object, an intuitive idea is to directly apply existing open-vocabulary object detection methods~\cite{regionclip,detic,vldet,glip} to solve the part detection/segmentation task. However, as shown in Table~\ref{table:motivation}, they do not show good generalization on part-level recognition. Although conceptually similar to open-vocabulary object detection, localizing and classifying the fine-grained object parts are essentially more challenging. This motivates us to explore new designs to empower current object detectors with open-vocabulary part segmentation ability.

\begin{table}[t]
\begin{center}
\input{tables/motivation.tex}
\end{center}
\vspace{-5mm}
\caption{\textbf{Performance of previous open-vocabulary object detection methods on Pascal Part~\cite{pascalpart} validation set.} The evaluation metric is mAP$_{\rm box}$@[.5, .95] on the detailed metrics of \texttt{dog}. All models use their official codebases and model weights. \gc{Oracle} is the method trained on Pascal Part training set.}
\label{table:motivation}
\end{table}

The model of open-vocabulary part segmentation is supposed to be able to segment the object not only on open category but also on open granularity. As shown in Figure~\ref{fig:teaser}, the [\texttt{dog}] can be parsed to the [\texttt{head, torso, leg, tail}], while in the finer granularity, the
head of a dog can be further parsed to the [\texttt{ear, eye, nose,} etc.]. Annotating such fine-grained object part is extremely expensive. Publicly available datasets of part segmentation are less rich and diverse than those of image classification and object detection datasets. Even though we collect three sources of part segmentation datasets, including Pascal Part~\cite{pascalpart}, PartImageNet~\cite{partimagenet}, and PACO~\cite{paco}, only a small number of objects part are accessible.

To expand the vocabulary of part categories, we first seek to utilize the large vocabulary object-level and image-level data, such as LVIS~\cite{lvis} and ImageNet~\cite{imagenet}, where object categories are known, but their part locations or part names are not. To enable part segmentation task benefit from them, our detector is based on the vision-language model~\cite{clip}, and trained on the joint of part-level, object-level and image-level data, where the classifier weight in the detector is replaced to the text embedding of the class name. In this way, the model learns to align vision and language at multi-granularity level to help generalize the ability to parse the object into its parts from base objects to novel objects.

Though the multi-granularity alignment is established, the part-level alignment for novel objects is fragile since its supervision signal is absent. To further strengthen it, we propose to leverage the pre-trained foundation models~\cite{dinoself} to parse the novel object into its parts as the annotations: 1) We find the nearest base object for each novel object by the similarity of their global features. 2) We build the dense semantic correspondence between the novel object and its corresponding base object by the similarity of their spatial features. 3) We parse the novel object into its parts in the way of the base object by the correspondence. The name of novel parts follows its corresponding base object. According to this pipeline, we generate the parsed images and use them as part annotations of novel objects.

Extensive experiments demonstrate that our method can significantly improve the open-vocabulary part segmentation performance. For cross-dataset generalization on PartImageNet, our method outperforms the baseline by 3.3$\sim$7.3 mAP. For cross-category generalization within Pascal Part, our approach improves the baseline by 7.3 AP$_{50}$ on novel parts. Finally, we train a detector with the joint data of LVIS, ImageNet, PACO, Pascal Part, PartImageNet, and parsed ImageNet. On three trained part segmentation datasets, it obtains better performance than their dataset-specific training. Meanwhile, part segmentation on a large range of objects in the open-world is achieved , as shown in Figure~\ref{fig:teaser}.

Our contributions are summarized as follows:
\begin{itemize}[noitemsep,nolistsep,leftmargin=*]
    \item We set up benchmarks and baseline models for open-vocabulary part segmentation in Pascal Part and PartImageNet datasets.   
    \item We propose a parsing pipeline to enable part segmentation to benefit from various data sources and expand the vocabulary of part categories.
    \item We train a detector with the ability of open-vocabulary object detection and part segmentation, achieving favorable performance on a wide range of part segmentation datasets.
\end{itemize}

\section{Related Work}

\paragraph{Open-vocabulary object detection.} OVOD~\cite{OVRCNN} aims to improve the generalization ability of object detectors from seen categories to novel categories. For example, ViLD~\cite{ViLD}, RegionCLIP~\cite{clip}, PB-OVD~\cite{gao2021towards} use pseudo region annotations generated from the pre-trained vision-language model~\cite{clip,li2021align}. DetPro~\cite{du2022learning} designs an automatic prompt learning method to improve the category embedding effectively. GLIP~\cite{glip} trains the detector on both detection and grounding data. Detic~\cite{detic} enlarges the number of novel classes with image classification data. VLDet~\cite{vldet} extracts region-word pairs from image-text pairs in an online way. Different from these works, we explore more fine-grained object recognition at the part level.

\paragraph{Part segmentation.} Beyond recognizing objects through category labels, a more fine-grained understanding of objects at the part level~\cite{de2021part,li2022ppart,zhou2021differentiable,michieli2020gmnet} is in increasing demand. Some pioneering works provide part annotations for specific domains, such as human~\cite{gong2017look, li2017multiple, yang2019parsing}, birds~\cite{wah2011caltech}, cars~\cite{reddy2018carfusion, song2019apollocar3d}, fashion domain~\cite{jia2020fashionpedia, zheng2018modanet}. Part annotations for common objects include such as Pascal-Part~\cite{pascalpart}, PartNet~\cite{partnet}, PartImageNet~\cite{partimagenet}, ADE20K~\cite{zhou2019semantic}, Cityscapes-Panoptic-Parts~\cite{meletis2020cityscapes} and more recent PACO~\cite{paco}. Based on these valuable datasets, our work is towards parsing any object in the open world.

\paragraph{Vision-and-language representation learning.} A universal representation for vision and language is needed in various tasks such as visual question answering~\cite{vqav2,gurari2018vizwiz}, image/video-text retrieval~\cite{coco,flickr30k,xu2016msr}, visual reasoning~\cite{suhr2017corpus,girdhar2019cater} and so on. To enhance the visual representation, region-based features obtained from object detection are introduced. For instance, OSCAR~\cite{li2020oscar} uses object tags and region features to train a universal semantic, UNITER~\cite{chen2020uniter} establishes word-region alignments by supervising similar output across multiple modalities, ALBEF~\cite{li2021align} aligns the image and text before fusing them with a multimodal encoder, SimVLM~\cite{wang2021simvlm} reduces the requirement to regional labels by exploiting large-scale weak supervision. Our work is aimed at learning part-level visual representation aligned with language supervision.

\paragraph{Semantic correspondence.} The aim of semantic correspondence ~\cite{han2017scnet,zhao2021multi,kim2017fcss,lee2019sfnet,min2019spair,yang2017object,liu2020semantic} is to establish the spatial visual correspondence between different instances of the same object category. Cross-domain correspondence~\cite{aberman2018neural, zhang2020cross} expands it to different categories. The pre-trained model is usually introduced to compute the feature map similarity. \cite{aberman2018neural} used a pre-trained CNN and \cite{dinodense} improved the performance by using ViT model~\cite{vit,dinodense}. In our work, we apply self-supervised DINO~\cite{dinoself} to build the semantic correspondence between the novel object and the base object.

\begin{figure}[t]
\begin{center}
\includegraphics[width=0.48\textwidth]{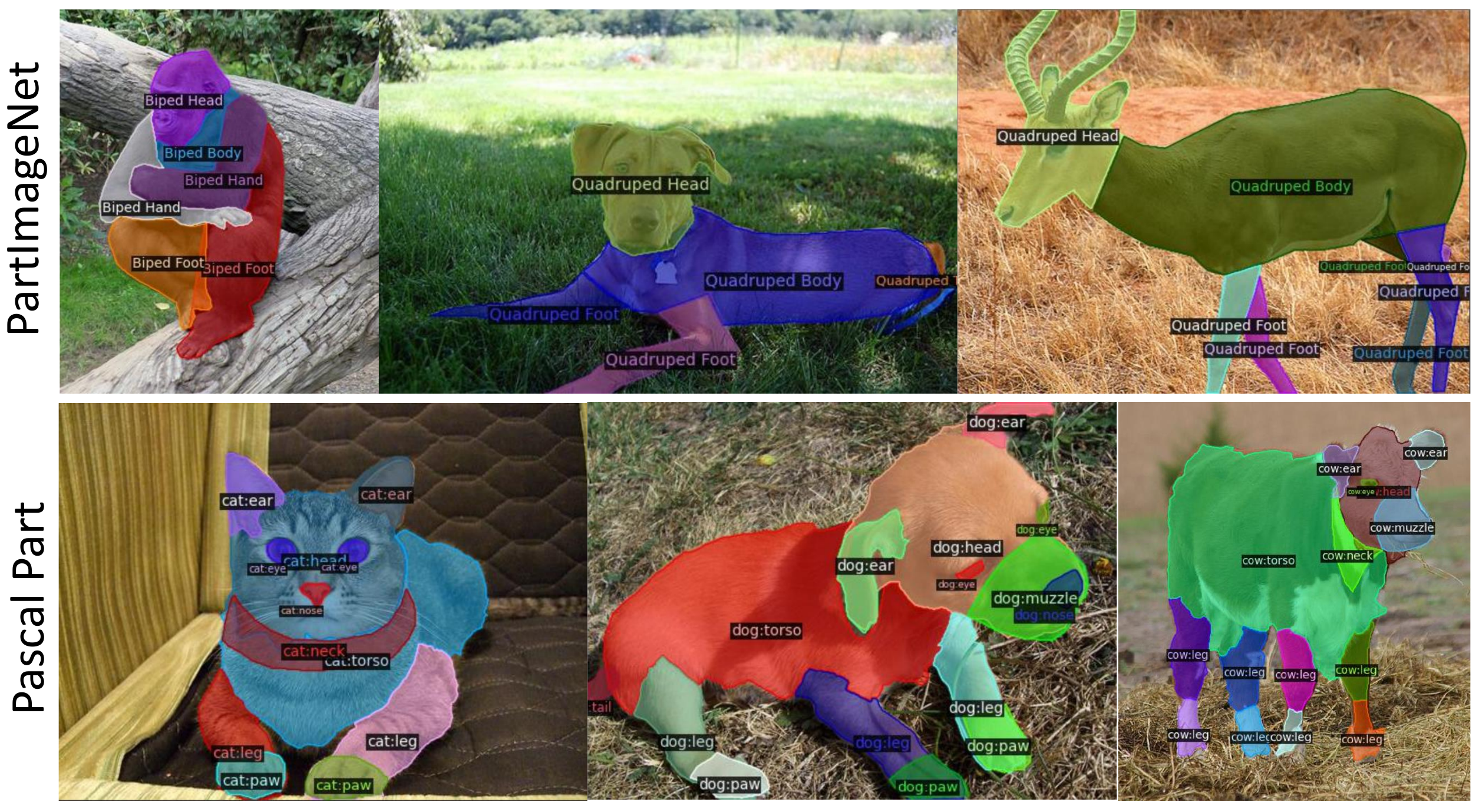}   
\end{center}
\vspace{-5mm}
\caption{\textbf{Different part taxonomies in different dataset annotations.} For example, the [\texttt{dog}] is parsed to the [\texttt{head, body, foot, tail}] in PartImageNet~\cite{partimagenet}. In the finer granularity, the head of a dog is further parsed to the [\texttt{ear, eye, muzzle, nose}] in Pascal Part~\cite{pascalpart}. }
\label{fig:taxonomy}
\end{figure}

\section{Open-Vocabulary Part Segmentation}
The goal of open-vocabulary part segmentation is to parse any object \textit{in the wild} into its components. The model takes as input the image and outputs its part segmentation by the pre-defined part taxonomy or custom text prompt.

\subsection{Open Category and Granularity}
We expect the model for open-vocabulary part segmentation to be able to provide the part segmentation in both open category and open granularity.

\paragraph{Open category.} Similar to object detection, open category means that the model is able to parse any category of the object \textit{in the wild}.

\paragraph{Open granularity.}
As shown in Figure~\ref{fig:taxonomy}, the part taxonomy is inherently hierarchical, the [\texttt{dog}] can be parsed to the [\texttt{head, body, leg, tail}], while in the finer granularity, the head of a dog can be further parsed to the [\texttt{ear, eye, muzzle, nose,} etc.]. This brings to inconsistent definitions in different datasets, for example, the taxonomy of Pascal Part~\cite{pascalpart} is relatively finer than  ImageNetPart~\cite{partimagenet}.

\subsection{Evaluation Protocol}
For terminology, \textit{base parts} are the parts of base objects (seen in the training set), and \textit{novel parts} are the parts of novel objects (unseen in the training set).

\paragraph{Cross-category generalization.}
 The models are trained on the base parts, such as parts of cat, cow, horse, and sheep, and evaluated on novel categories, such as parts of dog. In this setting, the base categories and the novel categories are \textit{mutually} exclusive.

\paragraph{Cross-dataset generalization.}
In practice use, the trained model could be used in any evaluation dataset, therefore, the object categories in inference may overlap with those in training. We use a cross-dataset setting to evaluate these more practical scenes.

\subsection{Revisit Open-Vocabulary Object Detection}
One may assume that the part instance is a special type of object, and the open-vocabulary part segmentation task can be solved by off-the-shelf open-vocabulary object detection/segmentation methods~\cite{regionclip,detic,vldet,glip}.

\paragraph{Adapting open-vocabulary object detector to part recognition.} To adapt the open-vocabulary object detector to part segmentation, its classifier weight in the region recognition head needs to be replaced by text embedding of the part category name. The performance of some popular open-vocabulary object detectors on part segmentation is shown in Table~\ref{table:motivation}. Obviously, their performances are far from satisfactory. We analyze the limited performance of open-vocabulary object detection on part segmentation comes from two folds: \textit{(i) Recall}. The part and the object have different granularities, it is non-trivial for region proposals trained on object-level data to generalize to part-level. \textit{(ii) Precision}. The learning materials of open-vocabulary object detectors, object-level and image-level data, have insufficient part instances and are hard to provide effective part-level supervision.

\begin{table}[t]
\begin{center}
\input{tables/recall.tex}
\end{center}
\vspace{-5mm}
\caption{\textbf{Evaluation on the generalizability of region proposals on objects and parts.} The recall is evaluated at IoU threshold 0.5 on the validation set of Pascal Part. All models are ResNet50 Mask R-CNN. The upper section is trained on object-level data and the lower section is part-level data. It is non-trivial for region proposals to generalize from object-level to part-level.}
\label{table:recall}
\vspace{-2mm}
\end{table}

\begin{figure}[t]
\begin{center}
\includegraphics[width=0.48\textwidth]{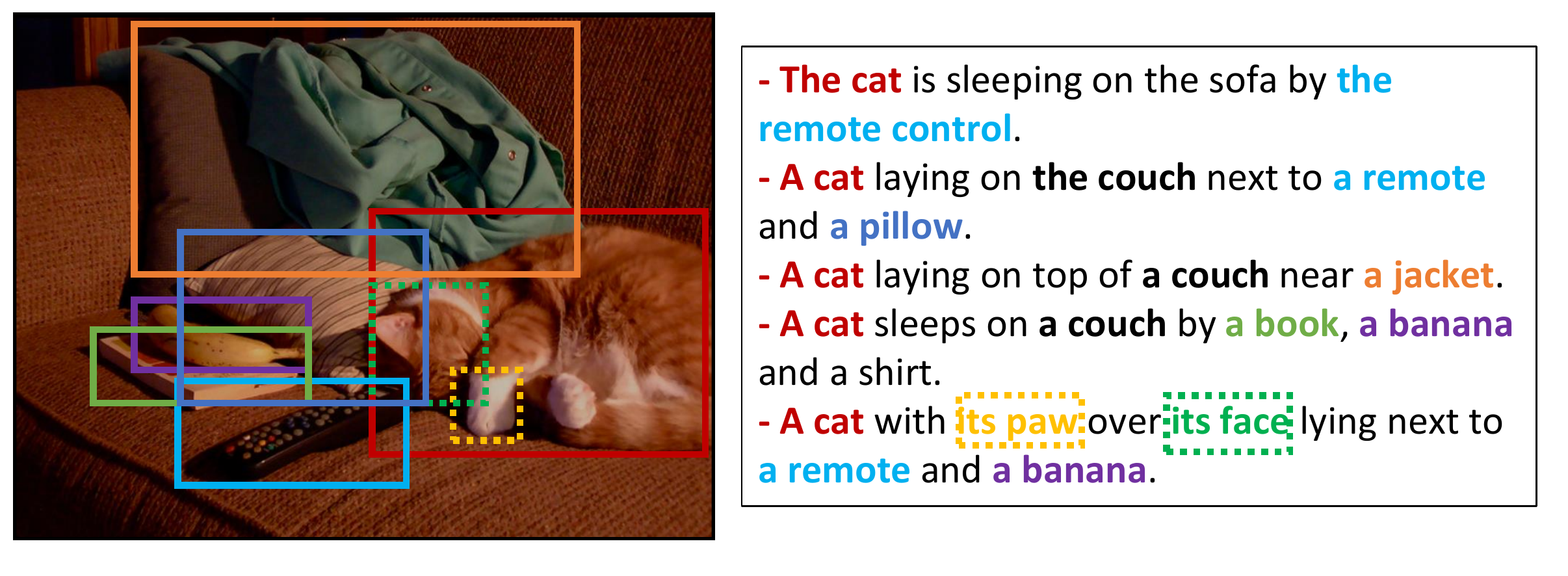}   
\end{center}
\vspace{-6mm}
\caption{\textbf{Example of COCO Caption~\cite{cococaption}.} COCO Caption data provides the image and its corresponding caption only, without object-level alignment (solid box) or part-level alignment (dashed box). Even if all alignments are known, part descriptions are much less frequent than object descriptions.
}
\label{fig:caption}
\end{figure}

\paragraph{Generalizability of region proposals.}
We study whether region proposal networks trained on object-level data could provide sufficient object proposals for the part. Although previous works~\cite{ViLD,detic} conclude that novel categories only suffer a small performance drop in recall, we point out that this is not the case when the novel objects have different granularity from object to part. As shown in Table~\ref{table:recall}, the detector trained on object-level data only has limited recall on the part dataset. To obtain better recall, part-level annotations are necessary, evidenced by the model trained on the Pascal Part base having very close recall with the fully-supervision model on the full Pascal Part training set.

\paragraph{Part-level alignment between image and its caption.}
Open-vocabulary object detection methods usually use image caption data to train the model. However, learning to detect the object part in the image from its image caption has two challenges: (1) Image caption data only provides the image and its corresponding caption, without dense captions on objects. Each open vocabulary object detector method~\cite{regionclip,detic,vldet,glip} needs to design its own method to align objects in the image and in the caption. (2) Even if the alignment could be extracted from the caption, or provided by the dataset annotations~\cite{plummer2015flickr30k,yu2016refcoco,krishna2017visualgenome}, we find the caption contains object parts less frequently than objects, as shown in Figure~\ref{fig:caption}. This less frequency makes part-level alignment between the image and its caption more difficult to learn than the object-level.

\section{Our Method}
\label{method}
Our detector architecture is a vision-language version of Mask R-CNN~\cite{maskrcnn}, where the classifier is the text embedding of category name from CLIP~\cite{clip}. This enables us to seamlessly train the detector on part-level, object-level and image-level data. We further parse the image data into its parts to expand the vocabulary of part categories, which is based on dense semantic correspondence between the base object and the novel object extracted from DINO~\cite{dinoself}.

\begin{figure*}[t]
\begin{center}
\includegraphics[width=1.0\textwidth]{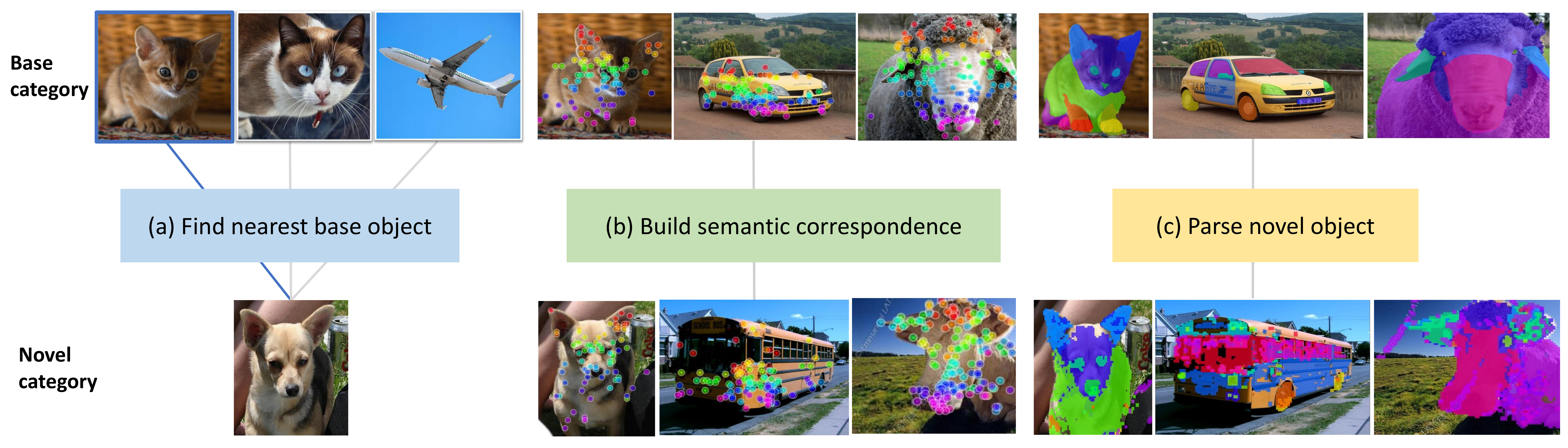}   
\end{center}
\vspace{-5mm}
\caption{\textbf{The pipeline of parsing novel objects into parts.} (a) Finding the nearest base object for each novel object. (b) Building the dense semantic correspondence between a novel object and its corresponding base object. For better visualization, we only some points sampled from the feature map grid. (c) Parsing the novel object as the way of the base object.}
\label{fig:pipeline}
\end{figure*}

\subsection{Detector Architecture}
\paragraph{Image encoder.} The image encoder is based on convolutional neural networks such as ResNet~\cite{resnet}  or Transformer-based models like Swin~\cite{swin}, followed by Feature Pyramid Network~\cite{fpn} to generate multi-scale feature maps to be used in the detection decoder.

\paragraph{Detection decoder.} The architecture of detection decoder is composed of a region proposal network (RPN)~\cite{FasterRCNN} and a R-CNN recognition head. RPN provides box proposals for both objects and parts. R-CNN recognition head refines the box location and the classification score. Notably, the classifier weight in the recognition head is replaced by text embedding of the class name of the object and the part.

\paragraph{Text embedding as the classifier.} The classification score of the recognition head is implemented as a dot-product operation between the region features and the text embeddings, where the region features are cropped from feature maps of the image encoder, and the text embeddings are extracted from the text encoder in CLIP~\cite{clip}.

\paragraph{Mask decoder.}
We choose the architecture of mask decoder from Mask R-CNN~\cite{maskrcnn} and replace the original multi-classification head with a class-agnostic head to support segmentation on novel categories. We note that more advanced architecture such as Mask2Former~\cite{mask2former} has the potential to further improve the performance but is not the focus of this work.

\subsection{Training on Parts, Objects, and Images}
The training data includes part-level, object-level, and image-level data. The image data is further parsed into the part annotation. Our detector is joint-trained on these data to establish multi-granularity alignment.

\paragraph{Part segmentation data.}
Part segmentation data~\cite{pascalpart,partimagenet,paco} contains part mask segmentation and its category. Part is always defined as an object-part pair since the same semantic part can be very different when it is associated with different objects. The category name of the part is formalized as follows:
\begin{equation}
\begin{aligned}
    C_{part} = [\text{``dog: head", ``dog: nose", ..., ``cat: tail"}] \nonumber
\end{aligned}
\end{equation}

\paragraph{Object detection data.} 
Object detection data contains object boxes and its category. Most object detection datasets~\cite{coco,lvis} also provide object mask segmentation annotations.
\begin{equation}
    C_{object} = \text{[``person", ``bicycle", ..., ``toothbrush"]} \nonumber
\end{equation}
The training loss for part and object data includes all location loss, classification loss, and mask loss.

\paragraph{Image classification data.}
Image classification data provides a large vocabulary of object categories in the form of images. Although object-level or part-level bounding annotations are absent, these images could be effectively used by the following ways: (1) The classification loss can be performed on max-size proposal~\cite{detic} for each image, and therefore expands the object-level vocabulary. (2) As will be introduced in section~\ref{parser}, the image can be parsed into parts and used as part-level annotations to expand the vocabulary of part categories. The training loss about image data only includes classification loss.

\subsection{Parsing Novel Objects into Parts}
\label{parser}
Most novel objects share the same part taxonomy with one of the base objects, for example, the novel \texttt{dog} has the same parts as the base  \texttt{cat}. Since the part segmentation of the base object is known, we could parse the novel object according to its dense semantic correspondence to the base object. The whole pipeline is shown in Figure~\ref{fig:pipeline}.

\paragraph{Finding the nearest base object for each novel object.} We use DINO~\cite{dinoself} to extract the [\texttt{class token}] of each base object, denoted as $t^{cls}(\cdot)$, and save these features as the database. Then, for each novel object $i$, we extract its feature using the same way and find its nearest base object $i_{near}$ in the database by the cosine similarity. 

\begin{equation}
\vspace{-1mm}
    i_{near} = \mathop{\arg\max}\limits_{j} \ {\rm sim}(t^{cls}(I_i), \ t^{cls}(I_j))
    \nonumber
\vspace{-1mm}
\end{equation}

\paragraph{Building dense semantic correspondence between the base object and its nearest novel object.}
We further use the DINO feature map as dense visual descriptors~\cite{dinodense}, denoted as $F_{x, y}
(\cdot)$, where $ x, y$ are grid indexes in the feature map. After computing the spatial similarity between the novel object $F_{x, y}(I_i)$ and its nearest base object $F_{p, q}(I_{i_{near}})$, for each token $(x, y)$ in the novel object, its corresponding token in the base object are chosen as the token with the highest cosine similarity.

\begin{equation}
\vspace{-1mm}
    x_{corr}, y_{corr} = \mathop{\arg\max}\limits_{p, q} \  {\rm sim}(F_{x, y}(I_i), \ F_{p, q}(I_{i_{near}}))
    \nonumber
\vspace{-1mm}
\end{equation}

\paragraph{Parsing novel parts by semantic correspondence.}
After dense correspondence between the base object and novel object is obtained, we could parse the novel object into its part segmentation $M_i(x, y)$ as the way of its corresponding base object part segmentation $M_{i_{near}}(p, q)$.
\begin{equation}
    M_i(x, y) = M_{i_{near}}(x_{corr}, y_{corr})
    \nonumber
\end{equation}

\paragraph{A hybrid parser to base and novel objects.}
Figure~\ref{fig:pipeline} also provides some examples of semantic correspondence and parsed novel objects. It can be seen that the strong and clear alignment is set up and the produced part segmentation is qualified to be used as pseudo part annotation for the novel object. For the base object, we use the detector trained on base parts to generate its pseudo part annotation.

\subsection{Inference on Text Prompt} 

In inference, the model takes as input the image and outputs the part segmentation for the object. Since all vocabulary of both objects and parts are a large number, and the user may not be interested in obtaining all possible object and part segmentation, our detector supports inference on text prompt by user input.

The left section of Figure~\ref{fig:teaser} is a case using a dog as an example. When the user-input is [\texttt{dog}], [\texttt{dog: head, torso, leg, tail}] and [\texttt{dog: head, ear, eye, nose, torso, leg, paw, tail}], the detector outputs the segmentation results in different granularities accordingly. The right section of Figure~\ref{fig:teaser} is a range of objects in the open world. It can be seen that our model is able to detect both open-vocabulary objects and their parts. When our detector is used in real applications, one can flexibly choose to use the pre-defined part taxonomy in datasets such as Pascal Part, PACO, or custom text prompt.

\section{Experiment}
\label{exp}

\subsection{Datasets}
We use three sources of part segmentation datasets, Pascal Part~\cite{pascalpart}, PartImageNet~\cite{partimagenet} and PACO~\cite{paco}.

\paragraph{Pascal Part.} The original Pascal Part provides part annotations of 20 Pascal VOC classes, a total of 193 part categories. Its taxonomy contains many positional descriptors, which is not suitable for this paper, and we modify its part taxonomy into 93 part categories.

\paragraph{PartImageNet.} PartImageNet groups 158 classes from ImageNet into 11 super-categories and provides their part annotations, a total of 40 part categories.

\paragraph{PACO.} PACO supplements more electronic equipment, appliances, accessories, and furniture than Pascal Part and PartImageNet. PACO contains 75 object categories, 456 object-part categories and 55 attributes. The image sources of PACO are LVIS and Ego4D~\cite{ego4d}. In this work, we use PACO-LVIS set as default. We focus on object parts and leave attributes for future research.

For object-level detection data, we use VOC~\cite{voc}, COCO~\cite{coco} and LVIS~\cite{lvis}. For image-level data, we use ImageNet1k (\textbf{IN})~\cite{imagenet}. We also create ImageNet-super11 (\textbf{IN-S11}) and ImageNet-super20 (\textbf{IN-S20}) that overlap with PartImageNet and Pascal category vocabulary separately. More details about datasets are in \Cref{sec:appendix-dataset}.

\subsection{Cross-dataset segmentation on PartImageNet}

\begin{table}[t]
    \begin{center}
        \begin{subtable}[ht]{0.98\linewidth}
        \input{tables/partimagenet.tex}
        \caption{\textbf{Cross-dataset generalization when only one part dataset, Pascal Part, is available.} Pascal Part is trained on the Pascal Part training set. IN-S11 label and Parsed IN-S11 are added into the training sequentially. }\label{table:partimagenet}
        \vspace{3mm}
        \end{subtable}
        \begin{subtable}[ht]{0.98\linewidth}
        \input{tables/partimagenet_data.tex}
        \caption{\textbf{Cross-dataset generalization when more than one part datasets are available.} Starting from Pascal Part, LVIS, PACO, IN-S11 and Parsed IN-S11 are added into the training sequentially.}\label{table:partimagenet_data}
        \end{subtable}
        \vspace{-2mm}
        \caption{\textbf{Cross-dataset generalization on PartImageNet part segmentation.} The evaluation metric is mAP$_{\rm mask}$@[.5, .95] on the validation set of PartImageNet. All models are ResNet50 Mask R-CNN and use the text embedding of the category name as the classifier. \gc{PartImageNet} is the fully-supervised method as the oracle performance.}\label{table:cross_dataset}  
    \end{center}
\vspace{-3mm}
\end{table}

In Table~\ref{table:cross_dataset}, we study cross-dataset generalization by using PartImageNet validation set as the evaluation dataset, where the metrics of all (40) parts and the detailed metrics of parts of \texttt{quadruped} are reported.

Table~\ref{table:partimagenet} shows when Pascal Part is the only available human-annotated part dataset, using IN-S11 data could help to improve PartImageNet performance.

\paragraph{Baseline from Pascal Part.} The baseline method directly uses the Pascal Part-trained model to evaluate PartImageNet. As shown in Table~\ref{table:partimagenet} first row, the performance is poor, for example, \texttt{body} and \texttt{foot} of the \texttt{quadruped} are nearly to zero. Pascal Part has no semantic label of \texttt{quadruped}, and the model needs to generalize from parts of \texttt{dog}, \texttt{cat}, etc. in Pascal Part to parts of \texttt{quadruped} in PartImageNet. The possible generalization ability comes from the text embedding generated from CLIP~\cite{clip}. However, generalization in part-level recognition is beyond its capability since CLIP is pre-trained on only image-level data.  

\paragraph{IN-S11 label.} Considering that Pascal Part has no semantic label such as \texttt{quadruped}, \texttt{piped}, \textit{etc.}, we collect IN-S11 images from ImageNet and add them to the training as image-level classification data. As shown in Table~\ref{table:partimagenet} second row, the performance is improved to some extent. This shows that image-level alignment is beneficial to the part recognition task. However, since no additional part-level supervision signal is introduced when using IN-S11 as image classification data, the improvement is still limited.

\paragraph{Parsed IN-S11.}
We use our parsing pipeline to deal with IN-S11 images and generate their part annotations. As shown in the third row in Table~\ref{table:partimagenet}, introducing these parsed parts into the training brings a significant improvement, 3.5$\sim$17.6 mAP improvement on the parts of \texttt{quadruped} and 3.3 mAP gain on all 40 parts over the baseline method. This suggests that our proposed methods are able to provide an effective part-level supervision signal to the detection model and boosts its performance on cross-dataset generalization.

\paragraph{More part datasets are available.}
Table~\ref{table:partimagenet_data} shows when more than one human-annotated part datasets are available, including Pascal Part, PACO, and LVIS. Although LVIS is an object-level dataset, we find its categories contain many object parts, such as shoes, which can also be seen as parts. From the first two rows of Table~\ref{table:partimagenet_data}, we can see that when the part-level annotations grow in training, the part segmentation obtains better performance, from 4.5 mAP to 7.8 mAP. When IN-S11 label and parsed IN-S11 are added to the training, the performance is further boosted by a large margin. For example, the head of \texttt{quadruped} has achieved 47.5 mAP, close to fully-supervised 57.3 mAP. This shows that when more data sources are available in the future, a strong model for part segmentation in the open world is promising.

\begin{table}[t]
\begin{center}
\input{tables/pascalpart.tex}
\end{center}
\vspace{-5mm}
\caption{\textbf{Cross-category generalization on Pascal Part part segmentation}. The evaluation metric is on the validation set of the Pascal Part. All models are ResNet50 Mask R-CNN and use the text embedding of the category name as the classifier. Base part is the base split from Pascal Part.
VOC object, IN-S20 label and Parsed IN-S20 are added into the training sequentially. \gc{Pascal Part} is the fully-supervised method as the oracle performance.} 
\label{table:pascalpart} 
\end{table}

\subsection{Cross-category segmentation on Pascal Part}
We evaluate the cross-category generalization within the Pascal Part dataset. All 93 parts are split into 77 base parts and 16 novel parts, detailed in Appendix. Table~\ref{table:pascalpart} reports the metrics of all (93), base (77), and novel (16) parts.

\paragraph{Baseline from Pascal Part base.} Table~\ref{table:pascalpart} first row is the baseline, which is trained on base parts and evaluated on novel parts. Since the detector uses CLIP text embedding as the classifier, the novel parts obtain non-zero segmentation performance.

\paragraph{VOC object.} 
Compared with the part annotation, the object annotation is much easier to collect. We add VOC object data to verify whether this could help to improve the performance. As shown in the second row of Table~\ref{table:pascalpart}, adding VOC object data helps to improve the performance on both base parts and novel parts in Pascal Part. This demonstrates that object-level alignment could lead to better part-level performance.

\paragraph{IN-S20 label.} Image-level classification data is also an easy-to-get annotation. We collect images with Pascal categories from ImageNet, IN-S20, and add them to the training. As shown in Table~\ref{table:pascalpart} third row, additional image-level data does not bring much gain than object detection data. This is because image-level data has a similar effect as object-level data on part-level recognition. Most of its gain is diminished by object data.

\paragraph{Parsed IN-S20.} We use our proposed parsing method to generate part annotations for novel objects, and they provide supervision on part classification. As shown in Table~\ref{table:pascalpart} fourth row, our method improves the performance on both base and novel categories. This shows that our parsing pipeline is an effective solution to both base and novel object part segmentation.

\begin{table}[t]
    \begin{center}
        \begin{subtable}[ht]{0.98\linewidth}
        \input{tables/across_r50.tex}
        \caption{All models are ResNet50~\cite{resnet} Mask R-CNN~\cite{maskrcnn}.}\label{table:across_r50}
        \vspace{3mm}
        \end{subtable}
        \begin{subtable}[ht]{0.98\linewidth}
        \input{tables/across_swinbase.tex}
        \caption{All models are Swin-B~\cite{swin} Cascade Mask R-CNN~\cite{CascadeRCNN}.}\label{table:across_swinbase}
        \end{subtable}
        \vspace{-2mm}
        \caption{\textbf{Part segmentation across datasets.} All models are evaluated by setting the classifier as text embedding of category name in the evaluation dataset. Joint denotes the joint-training on LVIS, PartImageNet, Pascal Part and PACO datasets. \gc{Dataset-specific} uses the training data of each dataset, separately.}\label{table:across}  
    \end{center}
\vspace{-3mm}
\end{table}

\subsection{Part segmentation across datasets}

Towards detecting and parse \textit{any} object in the open world, we train a detector on the joint of available part segmentation datasets, including LVIS, PACO, Pascal Part and PartImageNet. The performance is shown in Table~\ref{table:across}.

This joint training model shows good generalization ability on various evaluation datasets, for example, Pascal Part obtains 22.6 mAP, better performance than its dataset-specific training. However, the potential problem lies in that joint training does not benefit all datasets, where PartImageNet and PACO decrease the performance a little. 

To make up for the performance loss, we add IN and Parsed IN into the training. It can be seen all datasets obtain the performance gain accordingly. When we scale up the model capability from ResNet50~\cite{resnet} to Swin-B~\cite{swin}, the detector achieves better performance than dataset-specific training on all Pascal Part, PartImageNet and PACO datasets.

\begin{table}[t]
\begin{center}
\input{tables/prompt.tex}
\end{center}
\vspace{-5mm}
\caption{\textbf{Text prompt template to object part.} We compare different templates of text prompt to object part in the fully-supervision setting of Pascal Part and PartImageNet.}
\label{table:prompt} 
\end{table}

\begin{table}[t]
\begin{center}
\input{tables/align.tex}
\end{center}
\vspace{-5mm}
\caption{\textbf{Comparisons of different aligning methods for novel parts.} The experiments are carried out on cross-dataset generalization from Pascal Part to PartImageNet. Fine-tuning from the baseline model, max-score, max-size and our method apply different designs to utilize image-level data to further improve part segmentation performance, where the former two are trained on part labels expanded from the image label.}
\label{table:align} 
\end{table}

\subsection{Ablation Study}

\paragraph{Text prompt template.}  Since the part is associated with the object category, we study how to design the text prompt template of (\texttt{object}, \texttt{part}) pair to the text encoder. We select two common expressions:  a [\texttt{object}] [\texttt{part}] and  [\texttt{part}] of a [\texttt{object}]. For example, [\texttt{dog}] and [\texttt{head}], these two expressions are [\texttt{a dog head}] and [\texttt{head of a dog}]. As shown in Table~\ref{table:prompt}, a [\texttt{object}] [\texttt{part}] behaves a little better than [\texttt{part}] of a [\texttt{object}] in  Pascal Part while not in PartImageNet. Which expression is a generally better usage of text prompt to the part needs to be verified on more datasets and we leave it for future research. In addition, more advanced prompt engineering for part segmentation is also an open problem.

\paragraph{Aligning method for novel parts.} 
We compare different aligning methods to use IN-S11 data to help part segmentation in PartImageNet. We select two popular designs in open-vocabulary object detection, max-score and max-size. Max-score is selecting the proposal that has the highest score of the target category as the matched proposal, used in~\cite{regionclip}. Max-size is selecting the proposal that has the maximum area among all proposals as the matched proposal to the target category, proposed in
~\cite{detic}. For each ImageNet image, its object category is known, and its part taxonomy can be inferred, these parts will be used as the target category in max-score and max-size methods.

- \textit{Max-score.} As shown in Table~\ref{table:align} second row, max-score helps to improve the performance a little over baseline. Fine-tuning from the baseline model, its selected highest score proposals contain efficient training samples, and these samples bring performance gain.

- \textit{Max-size.} 
As shown in Table~\ref{table:align} third row, the max-size method degenerates the performance in most metrics. According to the max-size rule, all parts are assigned to the same proposal, it is hard to align the part-level region to its part name text embedding. This shows that part-level alignment is more difficult than object-level and an efficient fine-grained supervision signal is necessary.

\begin{figure}[t]
\begin{center}
\includegraphics[width=0.48\textwidth]{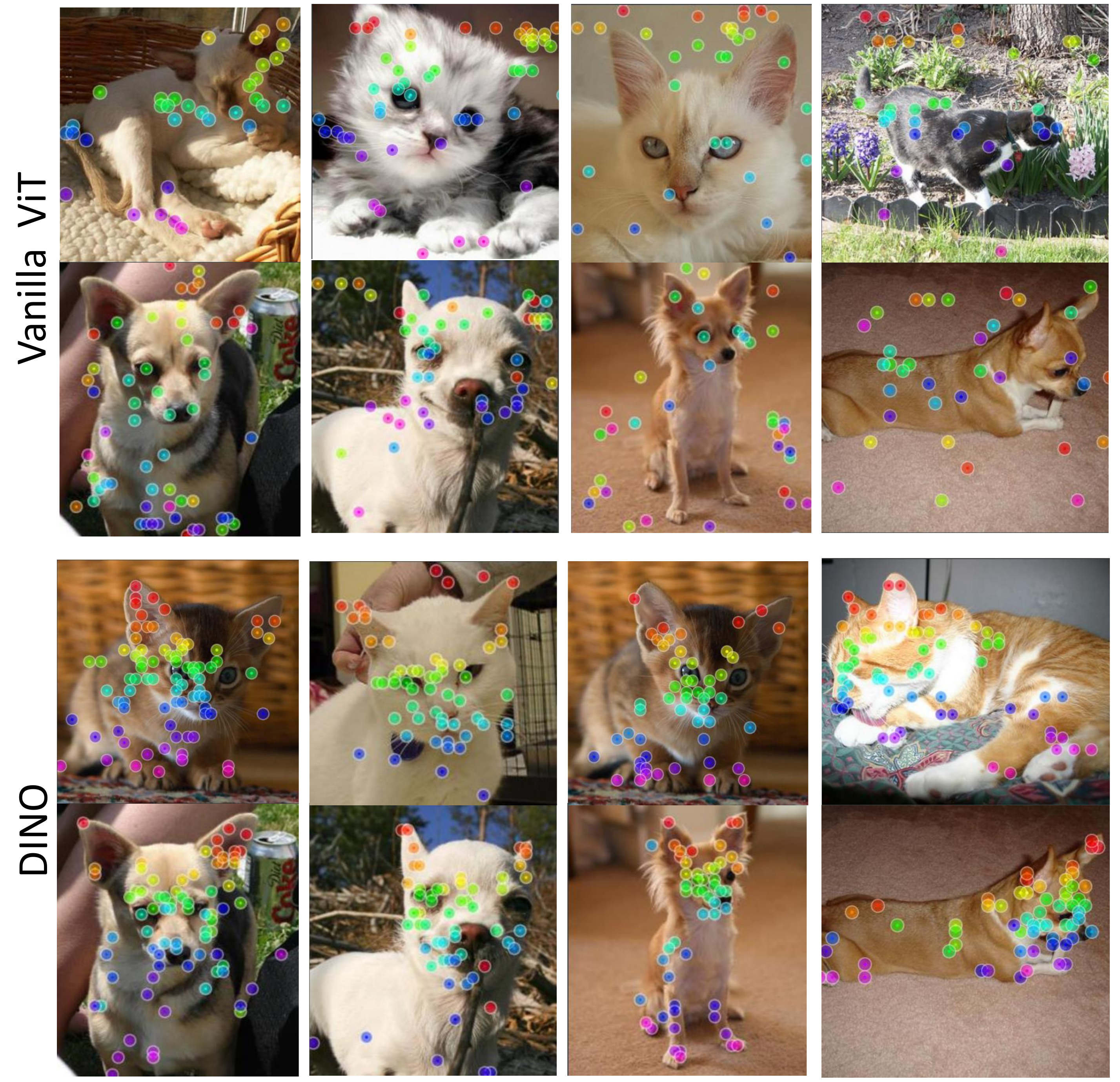}   
\end{center}
\vspace{-5mm}
\caption{\textbf{Semantic correspondence from vanilla ViT and DINO.} The upper section is from supervised ViT model~\cite{vit} and the lower section is from self-supervised DINO~\cite{dinoself}. For each section, first row and second row are paired base objects and novel objects. We crop each image into a uniform size for better visualization.}
\label{fig:vit}
\end{figure}

\paragraph{Pre-trained model in semantic correspondence.} We use self-supervised DINO~\cite{dinoself} in this work to find the nearest base object for each novel object and build their dense semantic correspondence. We verify whether a vanilla ViT~\cite{vit} has a similar function, which is pre-trained on fully-supervised ImageNet. As shown in Figure~\ref{fig:vit}, vanilla ViT is obviously behind DINO in the aspect of providing semantic correspondence. On the one hand, the nearest base object found by DINO has better alignment with the novel object in color, texture, and pose. On the other hand, the dense correspondence from DINO has clear semantics correspondence between the two objects. Similar experiment phenomenons are reported in ~\cite{dinoself,dinodense}. Besides DINO, whether other models could benefit to part segmentation is a potential research direction in the future. 

\begin{table}[t]
\begin{center}
\input{tables/foundation.tex}
\end{center}
\vspace{-5mm}
\caption{\textbf{The source of VLPart capability.} Besides training data of base parts and novel objects, two foundation models, CLIP and DINO, contribute to open-vocabulary part segmentation.}
\label{tab:foundation}
\end{table}

\section{Discussion}

\paragraph{Learning from Foundation Models}. When we analyze how VLPart achieves open-vocabulary part segmentation capability, as shown in Table~\ref{tab:foundation}, we could see that the important components of VLPart's capability are two foundation models: CLIP~\cite{clip} and DINO~\cite{dinoself}. Learning from foundation models is a recently rising research topic~\cite{bommasani2021opportunities,driess2023palm,alayrac2022flamingo,li2023blip}. Although a single foundation model is not an expert in a specific task, for example, neither CLIP nor DINO can accomplish the part segmentation task, combining these foundation models could bring to a range of applications~\cite{wu2023visual,liang2023taskmatrix,zhang2023prompt,liu2023visual,gong2023multimodal,2023interngpt,zhao2022learning,jatavallabhula2023conceptfusion}, and this paper takes the part segmentation task as an example to explore. In the future, how to "decode" more capabilities from foundation models is a very promising research topic.

\begin{figure}[t]
\begin{center}
\includegraphics[width=0.48\textwidth]{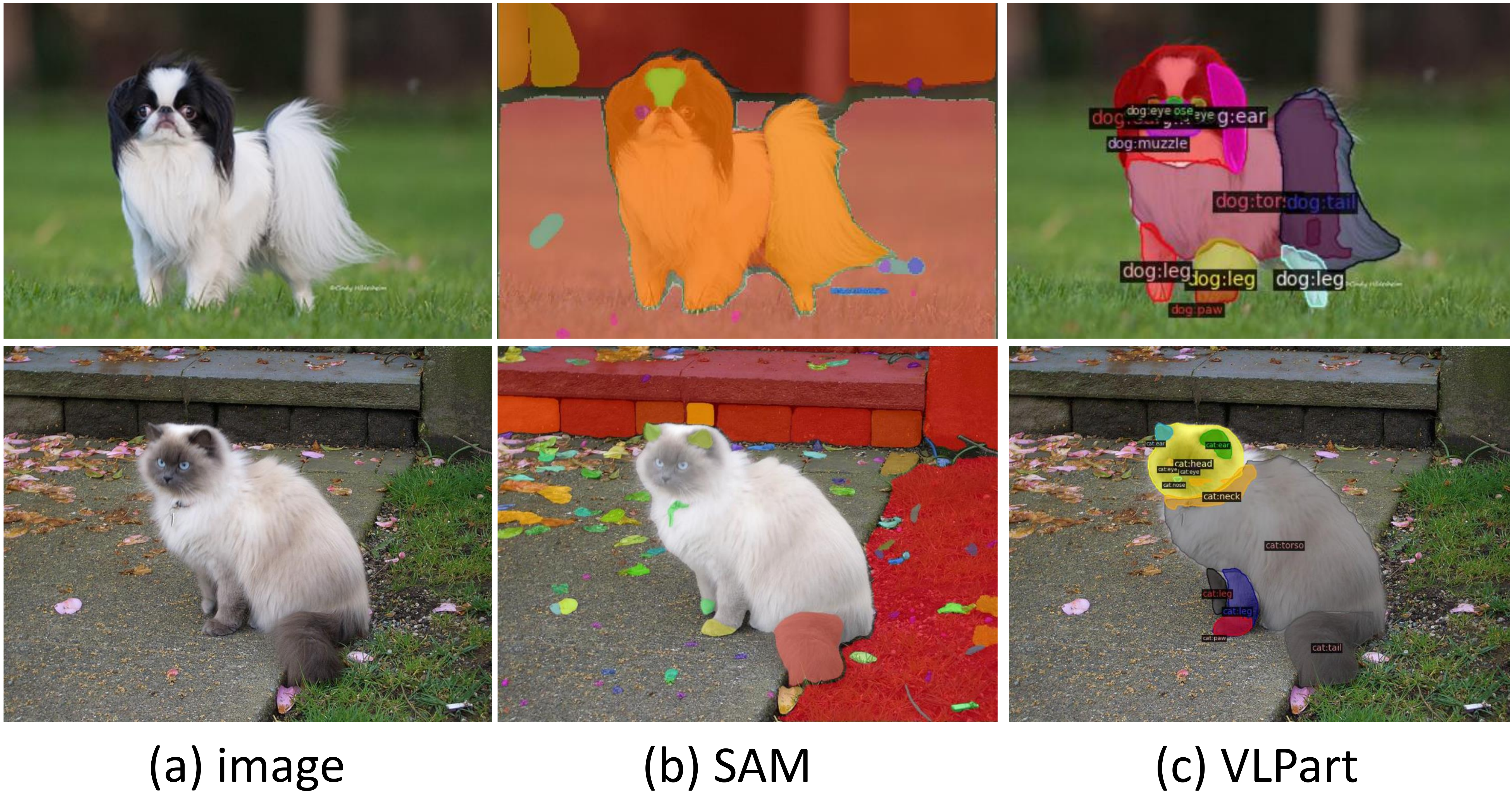}   
\end{center}
\vspace{-5mm}
\caption{\textbf{Comparsion of SAM~\cite{sam} and VLPart.} The main differences are: (1) SAM is a class-agnostic segmentation model and VLPart is class-aware, (2) SAM parses the object mostly in an edge-oriented way and VLPart is semantic-oriented.}
\label{fig:sam}
\end{figure}

\paragraph{Comparison with Segment Anything Model.}
Segment Anything Model (SAM)~\cite{sam} is a recently proposed model aimed to generate masks for all entities~\cite{qi2022open,qi2022fine} in an image, including both objects and their parts. As shown in Figure~\ref{fig:sam}, the main differences between SAM and VLPart are: (1) SAM is a class-agnostic mask segmentation model, while VLPart is class-aware. (2) The part segmentation of SAM is mostly edge-oriented, which makes it hard to parse two parts if there is no obvious edge between them, while VLPart parses objects based on semantics instead of low-level edge signals.

\paragraph{Segment and Recognize Anything Model?} A big picture for the vision perception system is to segment and recognize anything in the open world, in which SAM, open-vocabulary object detection and our open-vocabulary part segmentation are all sub-tasks. Some recently public works~\cite{liu2023grounding,zou2023segment,wang2023seggpt,yang2022psg} attempt to achieve this goal but their focuses are either segmentation or recognition. Furthermore, their explorations only reach the object-level, and do not go denser into the part-level. This paper provides a promising solution to part-level segmentation and recognition, serving as a component of achieving the goal of Segment and Recognize Anything Model.

\section{Conclusion and Future Work}
In this paper, we explore empowering object detectors with the fine-grained recognition ability of open-vocabulary part segmentation. Our model a vision-language version of the segmentation model to support text prompt input. The training data is the joint of part-level, object-level and image-level data to establish multi-granularity alignment. To further improve the part recognition ability, we parse the novel object into its parts by the dense semantic correspondence to its nearest base objects. Extensive experiments demonstrate that our method can
significantly improve the open-vocabulary part segmentation performance and achieve favorable performance on a wide range of datasets.

In the future, our models have great potential to be applied to various applications such as robotic manipulation~\cite{ge2022self}, part-guided instance object~\cite{paco}, and part-aware image editing~\cite{li2021partgan}.

\vspace{3mm}
\paragraph{Acknowledgments.} 
This work was done when Peize Sun worked as an intern at Meta AI and was supported in part by the General Research Fund of HK No.17200622.

\newpage

{\small
\bibliographystyle{ieee_fullname}
\bibliography{egbib}
}

\appendix

\section{Dataset}\label{sec:appendix-dataset}
In Table~\ref{tab:stat}, we list the datasets used in this paper.

\paragraph{Pascal VOC.} Pascal VOC contains 20 categories of objects and Pascal Part is annotated based on the same image set and object categories.

\paragraph{COCO.} COCO contains 80 categories of objects and is one of the most popular object detection datasets.

\paragraph{LVIS.} LVIS contains 1203 categories of objects and split them into frequent, common and rare sets according to the occurrence frequency. It is used to evaluate the long-tailed detection ability.

\begin{table}
\begin{center}
\input{tables_app/stat.tex}
\end{center}
\vspace{-5mm}
\caption{\textbf{Statistic of datasets used in this work.}}
\label{tab:stat}
\end{table}

\paragraph{Pascal Part.} The original Pascal Part is too much fine-grained, in Table~\ref{tab:ori_Pascalpart_taxonomy}, we modify it by (1) merging positional concepts, for example, merging the left wing and the right wing to the wing for the aeroplane, (2) deleting vague concepts, such as *side, (3) deleting train, tvmonitor. The resulting part taxonomy is 93 categories, as shown in Table~\ref{tab:Pascalpart_taxonomy}. We select \texttt{bus} and \texttt{dog} as the novel objects and their parts as the novel parts, totally 16 parts, the remaining 77 parts as base parts.

\paragraph{PartImageNet.}
PartImageNet selects 158 classes from ImageNet and groups them into 11 super-categories. For example in Table~\ref{tab:animal_quadruped}, \texttt{quadruped} super-category contains many animal categories. The part annotations are based on these super-categories, as shown in Table~\ref{tab:partimagenet_taxonomy}.

\paragraph{PACO.} PACO contains 75 object categories and 456 object-part categories. It supplements more electronic equipment, appliances, accessories, and furniture than Pascal Part and PartImageNet, as shown in Table~\ref{tab:pacopart_taxonomy}.

\begin{table}
\begin{center}
\input{tables_app/Pascalpart_list.tex}
\end{center}
\vspace{-6mm}
\caption{\textbf{Modified Pascal Part part taxonomy}.}
\label{tab:Pascalpart_taxonomy}
\vspace{4mm}
\end{table}

\begin{table}
\begin{center}
\input{tables_app/partimagenet_list.tex}
\end{center}
\vspace{-6mm}
\caption{\textbf{PartImageNet part taxonomy} from ~\cite{partimagenet}.}
\label{tab:partimagenet_taxonomy}
\vspace{4mm}
\end{table}

\begin{table}[t]
\begin{center}
\input{tables_app/partimagenet_quadruped.tex}
\end{center}
\vspace{-6mm}
\caption{\textbf{An example of PartImageNet super-category.} In PartImageNet annotation, only \texttt{quadruped} label is known without specific animal categories. }
\label{tab:animal_quadruped}
\vspace{4mm}
\end{table}

\paragraph{ImageNet.} ImageNet is the first large-scale image classification dataset. It has 1k version and 22k version. We use ImageNet-1k set as the default. ImageNet-super11 and -super20 are ImageNet images whose categories overlap with PartImageNet and Pascal category vocabulary, separately.

\begin{table*}[t]
    \begin{center}
        \begin{subtable}[ht]{0.98\linewidth}
        \input{tables_app/ovod_r50.tex}
        \caption{All models are ResNet50 Mask R-CNN.}\label{table:ovod_r50}
        \vspace{3mm}
        \end{subtable}
        \begin{subtable}[ht]{0.98\linewidth}
        \input{tables_app/ovod_swinbase.tex}
        \caption{All models are SwinBase Cascade Mask R-CNN.}\label{table:ovod_swin}
        \end{subtable}
        \vspace{-2mm}
        \caption{\textbf{Joint open-vocabulary object detection and part segmentation across datasets.} All models are evaluated by setting the classifier as text embedding of category name in the evaluation dataset. COCO is evaluated on object box detection, LVIS is on object mask segmentation, PartImageNet, Pascal Part and PACO are on part mask segmentation. \gc{Oracle} uses dataset-specific training data.}\label{tab:ovod}  
    \end{center}
\vspace{-3mm}
\end{table*}

\section{Joint object detection and part segmentation}

We evaluate on both object detection and part segmentation datasets, including VOC, COCO, LVIS, PartImageNet, Pascal Part and PACO. The performance is shown in Table~\ref{tab:ovod}. The joint training model shows good generalization ability on various datasets. For example, ResNet50 Mask R-CNN trained on LVIS, PACO and Pascal Part achieves zero-shot 28.6 AP on COCO and 7.8 AP on PartImageNet. However, the potential problem lies in that joint object detection and part segmentation do not benefit to both tasks all the time. For example, Pascal Part obtains better performance than its dataset-specific oracle, while LVIS decreases its performance. How to make these two tasks benefit from each other is a valuable question for future research.

\section{Part-level image editing}
\begin{figure}[ht]
    \centering
    \begin{subfigure}{\linewidth}
    \includegraphics[width=0.98\textwidth]{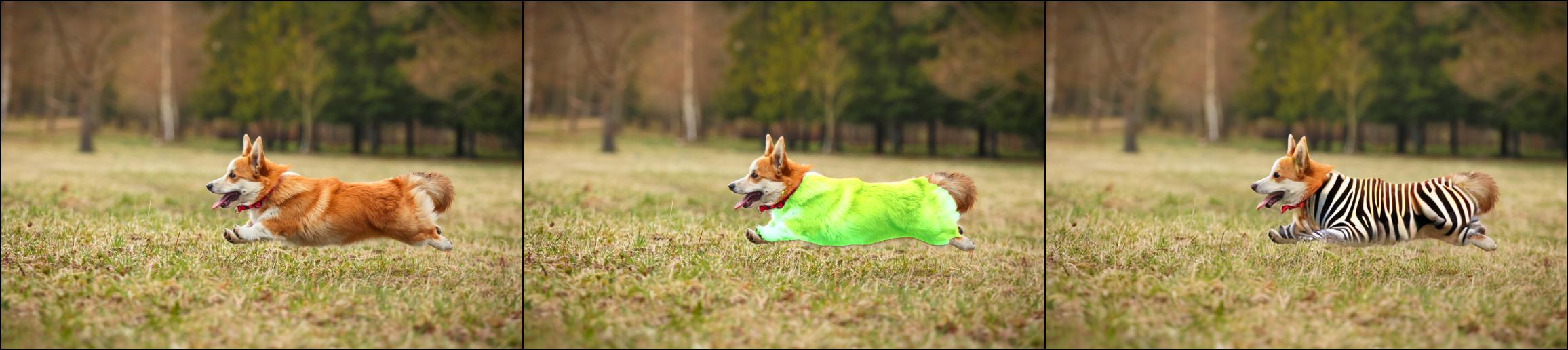}
    \caption{\texttt{dog body} $\rightarrow$ \texttt{zebra}}
    \end{subfigure}
    \begin{subfigure}{\linewidth}
    \includegraphics[width=0.98\textwidth]{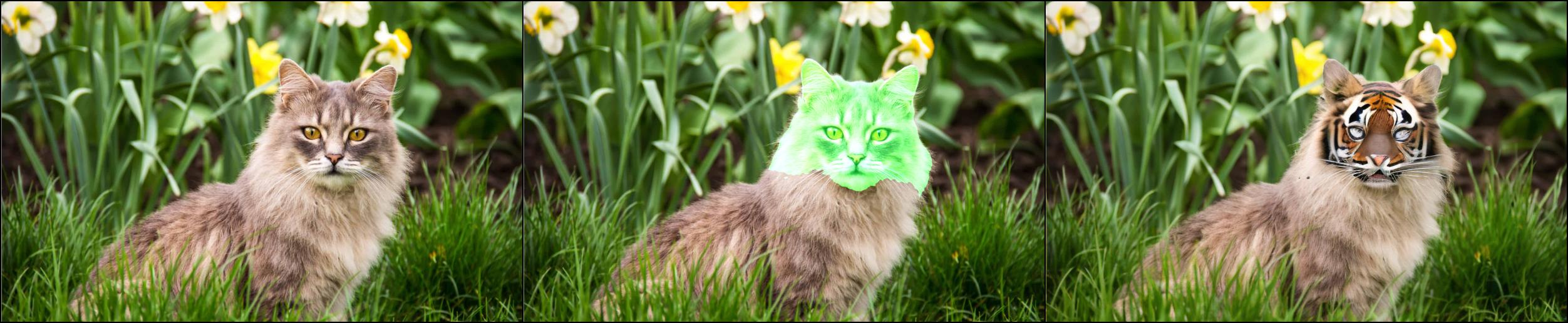}
    \caption{\texttt{cat head} $\rightarrow$ \texttt{tiger}}
    \end{subfigure}
    \begin{subfigure}{\linewidth}
    \includegraphics[width=0.98\textwidth]{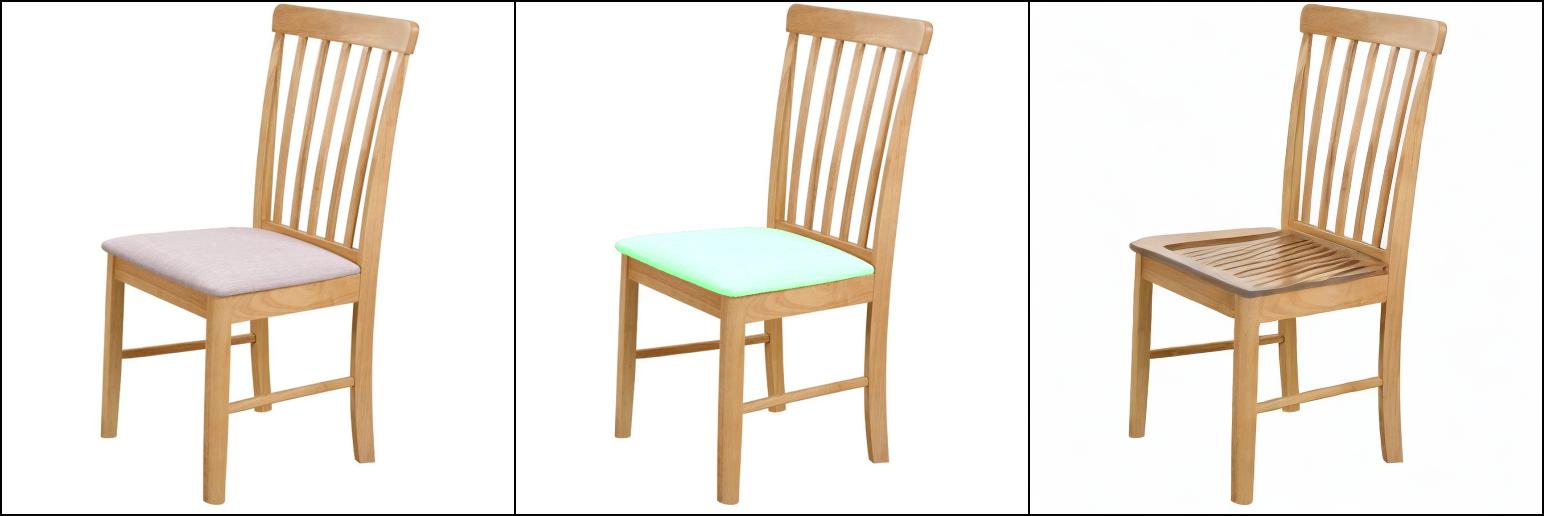}
    \caption{\texttt{chair seat} $\rightarrow$ \texttt{chocolate bar}}
    \end{subfigure}
    \caption{\textbf{Part-level image in-painting.} We first adopt VLPart to obtain text-prompt mask, and then use Stable Diffusion model to accomplish zero-shot in-painting inference.}
    \label{fig:part_inpaint}
    \vspace{-3mm}
\end{figure}

Given the part-level segmentation mask, we utilize Stable Diffusion~\cite{rombach2022high}, a state-of-the-art text-to-image generation model which performs well for in-painting tasks, to accomplish part-level image editing. Specifically, we first adopt VLPart to obtain text-prompt mask in the image, and then use a specifically fine-tuned version of Stable Diffusion model for image in-painting, to accomplish zero-shot in-painting inference.
Several examples of part-level in-painting are shown in \Cref{fig:part_inpaint}.

\section{A dialogue system}

\begin{figure*}[t]
\includegraphics[width=0.98\linewidth]{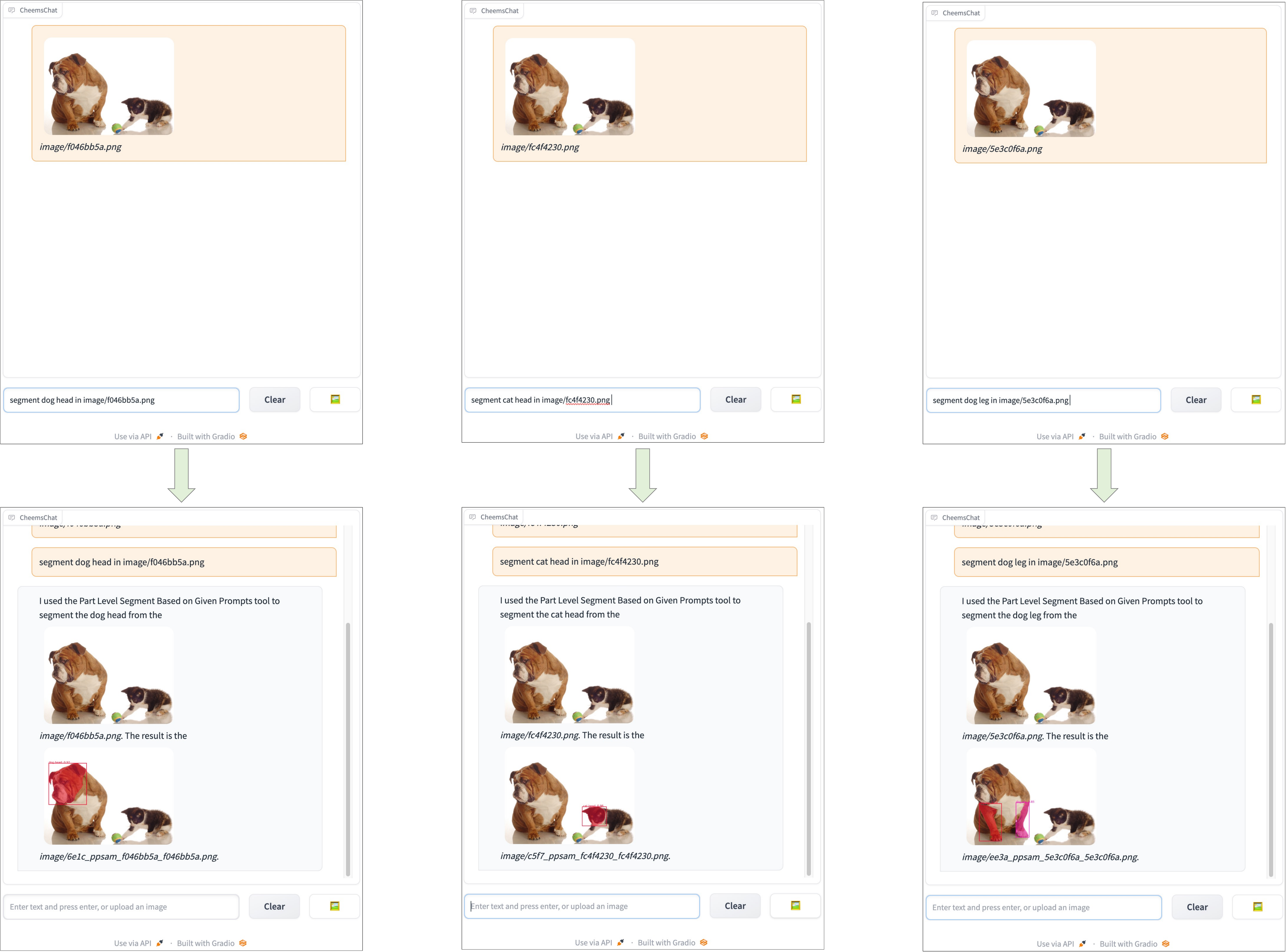}
\caption{\textbf{Demonstration video snapshots of a dialogue system} from \url{https://cheems-seminar.github.io/}.}
\label{fig:chatgpt}
\end{figure*}

Inspired by the design of Visual ChatGPT~\cite{wu2023visual}, we further build a multi-modal conversational system that interacts with VLPart by GPT-3~\cite{gpt3}~(\texttt{text-davinci-003}) and langchain~\cite{Chase_LangChain_2022}. The demonstration video snapshots are shown in \Cref{fig:chatgpt}.

\section{Visualization}

\begin{figure*}[t]
\begin{subfigure}{1.00\textwidth}
\begin{center}
\includegraphics[width=0.85\linewidth]{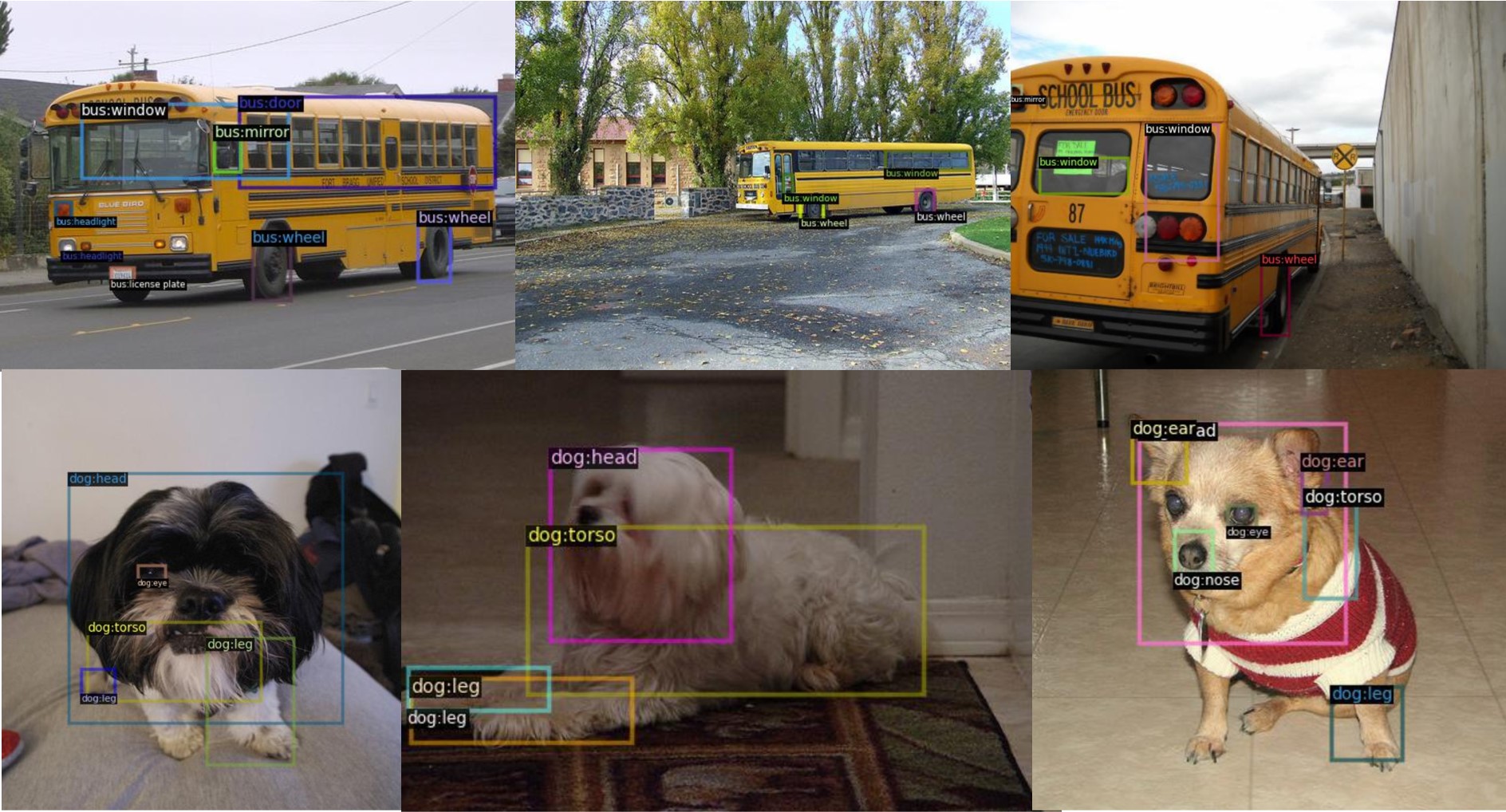}  
\end{center}
\caption{Parsed objects on Pascal images.}
\label{fig:parse1}
\vspace{5mm}
\end{subfigure}
\begin{subfigure}{1.00\textwidth}
\begin{center}
\includegraphics[width=0.85\linewidth]{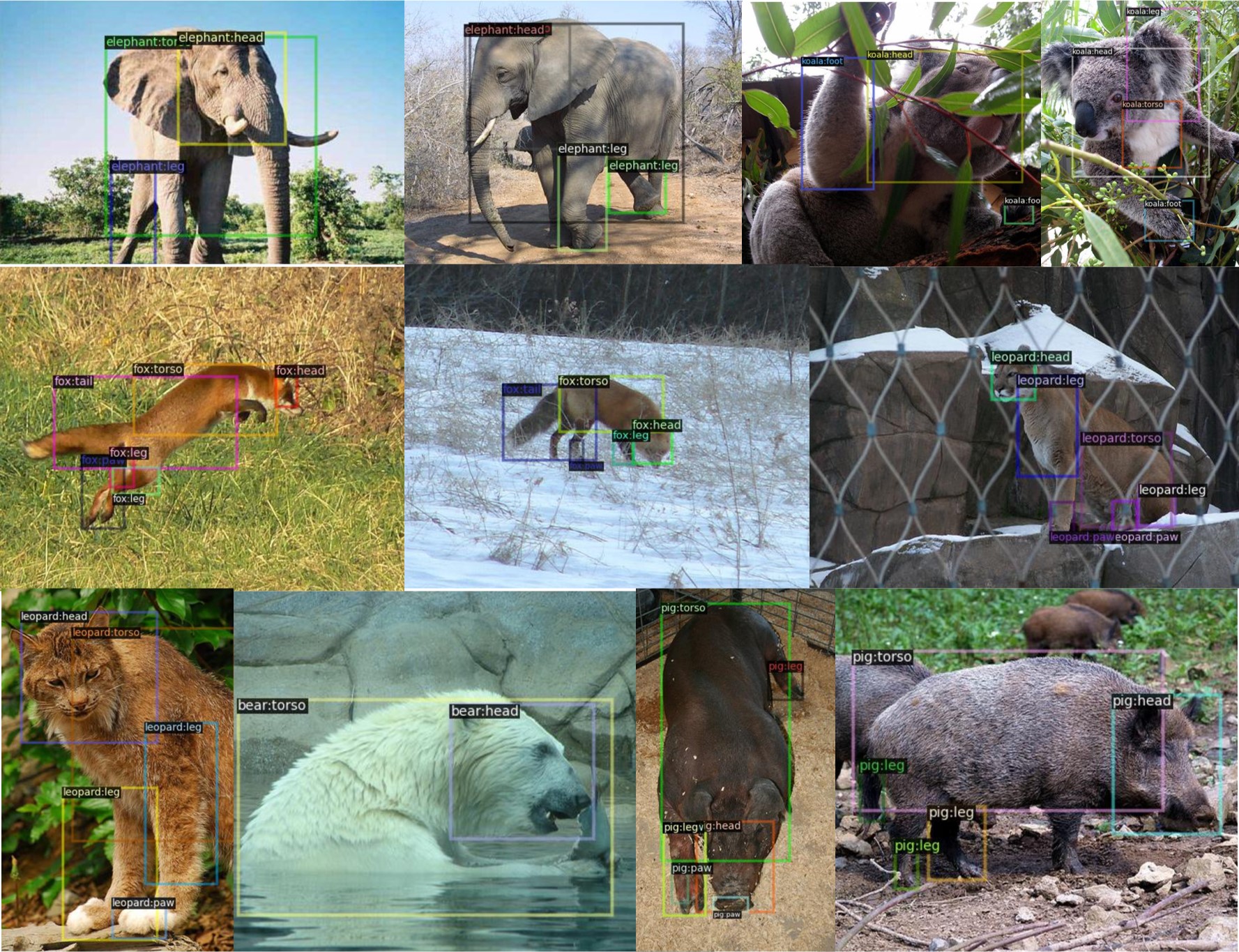}
\end{center}
\caption{Parsed objects on ImageNet images.}
\label{fig:parse2}
\end{subfigure}
\caption{\textbf{Visualization of parsing novel objects into parts}. Since we use the parsed image as the annotation to train the classification loss only, we only save the box and the category to the annotation files.}
\label{fig:parse}
\end{figure*}

\begin{figure*}[t]
\begin{subfigure}{0.95\textwidth}
\includegraphics[width=1.05\linewidth]{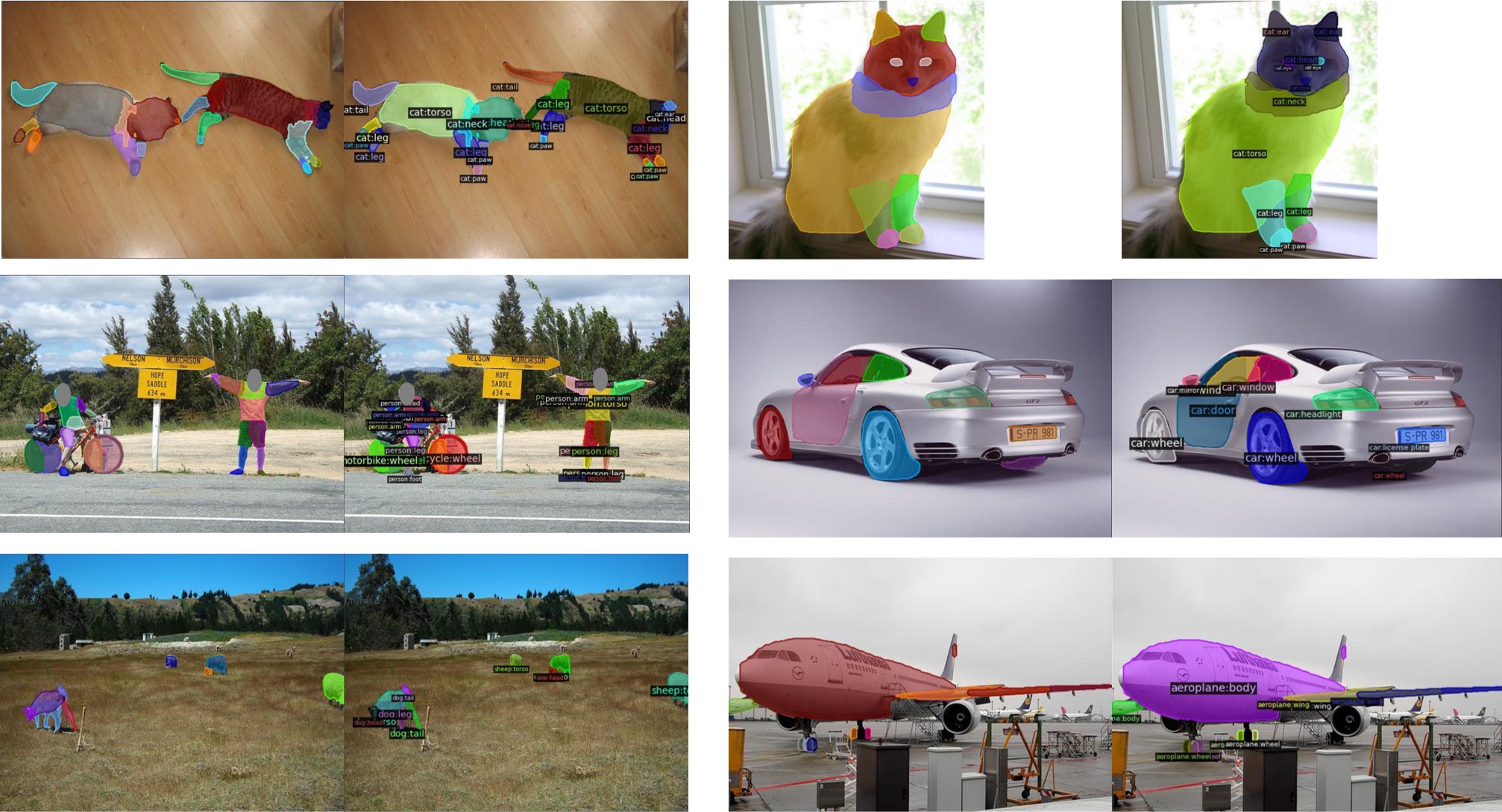}
\caption{Part segmentation on Pascal images using Pascal Part taxonomy.}
\label{fig:vis1}
\vspace{5mm}
\end{subfigure}
\begin{subfigure}{0.95\textwidth}
\includegraphics[width=1.05\linewidth]{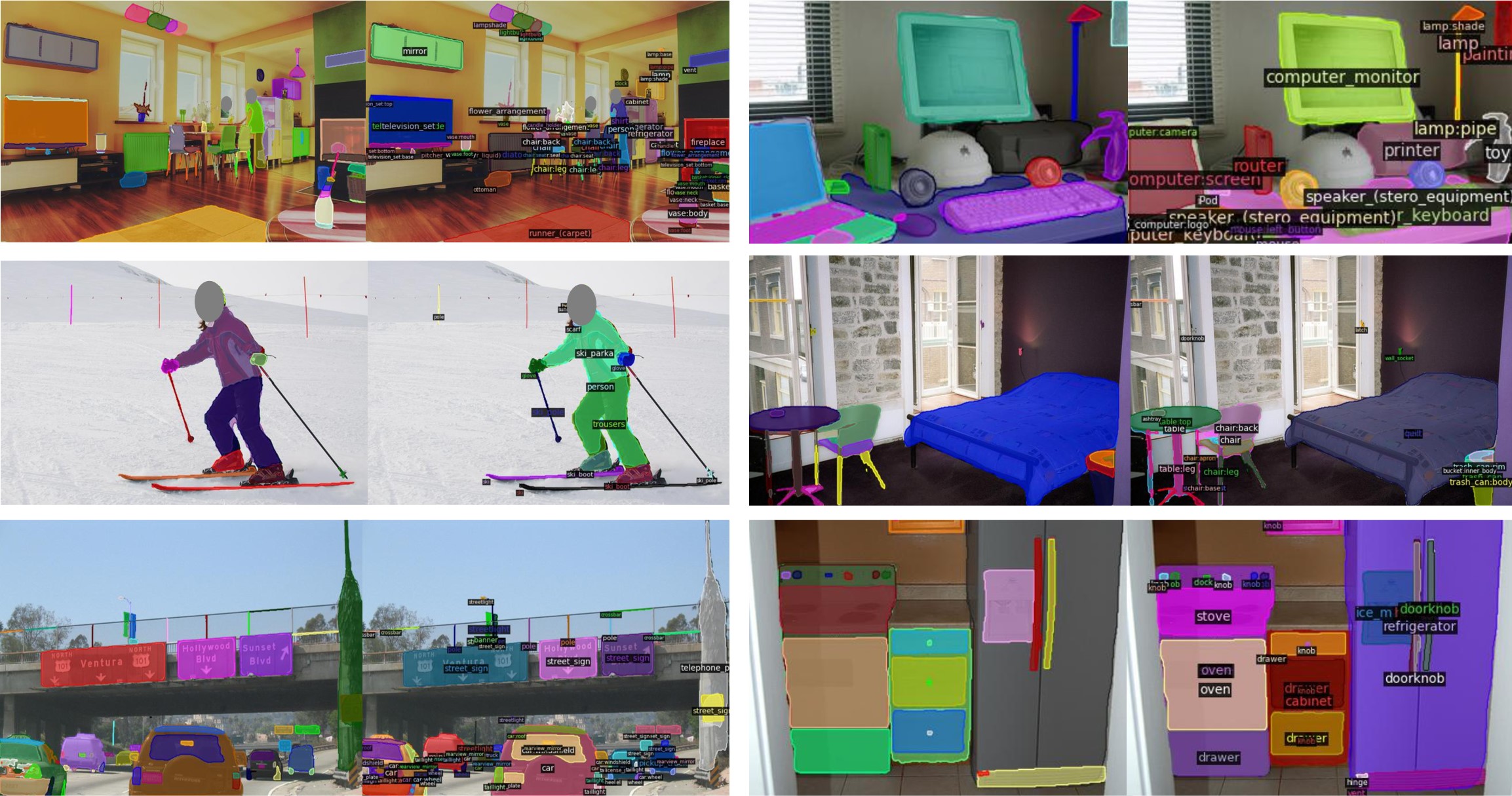}
\caption{Part segmentation on COCO images using joint taxonomy of LVIS and PACO.}
\label{fig:vis2}
\end{subfigure}
\caption{\textbf{Visualization of open-vocabulary object detection and part segmentation}. For each pair of images, left image is without category name for better visualization, right image is marked with category name.}
\label{fig:vis}
\end{figure*}

In Figure~\ref{fig:parse}, we show the parsed novel objects in Pascal and ImageNet images. In Figure~\ref{fig:vis}, we demonstrate the qualitative results of open-vocabulary part segmentation on Pascal and COCO images.

\begin{table*}
\begin{center}
\input{tables_app/Pascalpart_list_ori.tex}
\end{center}
\caption{\textbf{Original Pascal Part part taxonomy} from \url{ http://roozbehm.info/Pascal-parts/Pascal-parts.html}.}
\label{tab:ori_Pascalpart_taxonomy}
\end{table*}

\begin{table*}
\begin{center}
\input{tables_app/paco_list.tex}
\end{center}
\caption{\textbf{PACO part taxonomy} from~\cite{paco}.}
\label{tab:pacopart_taxonomy}
\end{table*}

\end{document}

%% file: tables/motivation.tex
\tablestyle{8.6pt}{1.2}
\begin{tabular}{lccccc}
\shline 
\multirow{2}{*}{Method} &
\multicolumn{5}{c}{dog} \\
 & head & leg & paw & tail & torso\\
\cmidrule(r){1-1}
\cmidrule(r){2-6}
RegionCLIP~\cite{regionclip} & 5.2 & 0.1 & 0.2 & 0.0 & 1.9\\
Detic~\cite{detic} & 3.2 & 0.0 & 0.0 & 0.0 & 2.0\\
VLDet~\cite{vldet} & 3.5 & 0.0 & 0.0 & 0.0 & 1.9\\
GLIP~\cite{glip} & 32.6 & 3.1 & 2.7 & 9.5 & 2.2 \\
\cmidrule(r){1-1}
\cmidrule(r){2-6}
\gc{Oracle} & \gc{50.7} & \gc{14.8} & \gc{20.7} & \gc{10.4} & \gc{18.7}\\
\shline
\end{tabular}

%% file: tables/recall.tex
\tablestyle{3.0pt}{1.2}
\begin{tabular}{l c c c c c }
\shline
\multirow{2}{*}{Training data} & 
\multirow{2}{*}{Type} &
\multicolumn{4}{c}{Pascal Part} \\
 & & AR@30 & AR@100 & AR@300 & AR@1000 \\
\cmidrule(r){1-2}
\cmidrule(r){3-6}
VOC & object & 7.7 & 11.8 & 15.4 & 16.1  \\
COCO & object & 8.4 & 14.8 &  24.4 & 40.5  \\
LVIS & object & 12.7 & 20.6 & 30.0 & 45.8  \\
\cmidrule(r){1-2}
\cmidrule(r){3-6}
Pascal Part base &  part & 29.4 & 48.1 & 63.6 & 75.3  \\
\gc{Pascal Part} &  \gc{part} & \gc{31.1} & \gc{50.5} & \gc{67.2} & \gc{78.8} \\
\shline
\end{tabular}

%% file: tables/partimagenet.tex
\tablestyle{8.5pt}{1.2}
\begin{tabular}{lccccc}
\shline
\multirow{2}{*}{Method} & 
All & 
\multicolumn{4}{c}{quadruped} \\
& (40) & head & body & foot & tail \\
\cmidrule(r){1-1}
\cmidrule(r){2-2}
\cmidrule(r){3-6}
Pascal Part & 4.5 & 17.4 & 0.1 & 0.0 & 2.9 \\
+ IN-S11 label & 5.4 & 23.6 & 3.4 & 0.8 & 1.2\\
+ Parsed IN-S11  & 7.8 & 35.0 & 15.2 & 3.5 & 8.9\\
\textit{vs. baseline}  & \greenc{+3.3} & \greenc{+17.6} & \greenc{+15.1} & \greenc{+3.5} & \greenc{+6.0}\\
\cmidrule(r){1-1}
\cmidrule(r){2-2}
\cmidrule(r){3-6}
\gc{PartImageNet} & \gc{29.7} & \gc{57.3} & \gc{25.8} & \gc{22.9} & \gc{22.9} \\
\shline
\end{tabular}

%% file: tables/partimagenet_data.tex
\tablestyle{8.5pt}{1.2}
\begin{tabular}{lccccc}
\shline
\multirow{2}{*}{Method} & 
All & 
\multicolumn{4}{c}{quadruped} \\
& (40) & head & body & foot & tail \\
\cmidrule(r){1-1}
\cmidrule(r){2-2}
\cmidrule(r){3-6}
Pascal Part & 4.5 & 17.4 & 0.1 & 0.0 & 2.9 \\
+ LVIS, PACO & 7.8 & 22.9 & 7.1 & 0.3 & 4.0\\
+ IN-S11 label & 8.8 & 26.3 & 3.7 & 0.4 & 1.0 \\
+ Parsed IN-S11 & 11.8 & 47.5 & 13.4 & 4.5 & 14.8\\
\textit{vs. baseline} & \greenc{+7.3} & \greenc{+30.1} & \greenc{+13.3} & \greenc{+4.5} & \greenc{+11.9} \\
\cmidrule(r){1-1}
\cmidrule(r){2-2}
\cmidrule(r){3-6}
\gc{PartImageNet} & \gc{29.7} & \gc{57.3} & \gc{25.8} & \gc{22.9} & \gc{22.9} \\
\shline
\end{tabular}

%% file: tables/pascalpart.tex
\tablestyle{6.0pt}{1.2}
\begin{tabular}{lcccccc}
\shline 
\multirow{2}{*}{Method} & 
\multicolumn{2}{c}{All (93)} &
\multicolumn{2}{c}{Base (77)} &
\multicolumn{2}{c}{Novel (16)} \\
& AP & AP$_{50}$ &
AP & AP$_{50}$ &
AP & AP$_{50}$  \\
\cmidrule(r){1-1}
\cmidrule(r){2-3}
\cmidrule(r){4-5}
\cmidrule(r){6-7}
Base part &  15.0 & 33.4 & 17.8 & 39.6	& 1.5 & 3.7 \\
+ VOC object & 16.8 & 36.8 & 19.9 & 43.3 & 2.1 & 5.9 \\
+ IN-S20 label & 17.4 & 37.5 & 20.8 & 44.7 & 1.1 & 3.1 \\
 + Parsed IN-S20 & 18.4 & 39.4	& 21.3 & 45.3 &	4.2 & 11.0 \\
\textit{vs. baseline}  & \greenc{+3.4} & \greenc{+6.0} & \greenc{+3.5} & \greenc{+5.7} & \greenc{+2.7} & \greenc{+7.3} \\
\cmidrule(r){1-1}
\cmidrule(r){2-3}
\cmidrule(r){4-5}
\cmidrule(r){6-7}
\gc{Pascal Part} &  \gc{19.4} & \gc{42.7} & \gc{18.8} & \gc{41.5} & \gc{22.1} & \gc{48.9}   \\
\shline
\end{tabular}

%% file: tables/across_r50.tex
\tablestyle{6.0pt}{1.2}
\begin{tabular}{lcccccc}
\shline
\multirow{2}{*}{Method} &  
\multicolumn{2}{c}{PartImageNet} &
\multicolumn{2}{c}{Pascal Part} &
\multicolumn{2}{c}{PACO}
\\
& AP & AP$_{50}$ & 
 AP & AP$_{50}$ & 
 AP & AP$_{50}$ 
\\
\cmidrule(r){1-1}
\cmidrule(r){2-3}
\cmidrule(r){4-5}
\cmidrule(r){6-7}
Joint & 29.1 & 52.0 & 22.6 & 47.8 & 9.3 & 18.9\\
+ IN & 30.8 & 54.4 & 23.6 & 49.2 & 9.0 & 18.7 \\
+ Parsed IN & 31.6 & 55.7 & 24.0 & 49.8 & 9.6 & 20.2\\
\textit{vs. baseline}  & \greenc{+2.5} & \greenc{+3.7} & \greenc{+1.4} & \greenc{+2.0} & \greenc{+0.3} & \greenc{+1.3} \\
\cmidrule(r){1-1}
\cmidrule(r){2-3}
\cmidrule(r){4-5}
\cmidrule(r){6-7}
\gc{Dataset-specific} & \gc{29.7} & \gc{54.1} & \gc{19.4} & \gc{42.3} & \gc{10.6} & \gc{21.7} \\
\shline
\end{tabular}

%% file: tables/across_swinbase.tex
\tablestyle{6.0pt}{1.2}
\begin{tabular}{lcccccc}
\shline
\multirow{2}{*}{Method} &  
\multicolumn{2}{c}{PartImageNet} &
\multicolumn{2}{c}{Pascal Part} &
\multicolumn{2}{c}{PACO}
\\
& AP & AP$_{50}$ & 
 AP & AP$_{50}$ & 
 AP & AP$_{50}$ 
\\
\cmidrule(r){1-1}
\cmidrule(r){2-3}
\cmidrule(r){4-5}
\cmidrule(r){6-7}
Joint & 40.0 & 64.8 & 31.2 & 60.5 & 15.4 & 30.3\\
+ IN & 41.2 & 66.8 & 31.7 & 61.1 & 15.9 & 30.8 \\
+ Parsed IN & 42.0 & 68.2 & 31.9 & 61.6 & 15.6 & 30.6\\
\textit{vs. baseline}  & \greenc{+2.0} & \greenc{+3.4} & \greenc{+0.7} & \greenc{+0.9} & \greenc{+0.2} & \greenc{+0.3} \\
\cmidrule(r){1-1}
\cmidrule(r){2-3}
\cmidrule(r){4-5}
\cmidrule(r){6-7}
\gc{Dataset-specific} & \gc{41.7} & \gc{68.7} & \gc{27.4} & \gc{56.1} & \gc{15.2} & \gc{29.4} \\
\shline
\end{tabular}

%% file: tables/prompt.tex
\tablestyle{7.2pt}{1.2}
\begin{tabular}{lccccc}
\shline
\multirow{2}{*}{Pascal Part} & 
All & 
\multicolumn{4}{c}{dog} \\
& (93) & head & torso & paw & tail \\
\cmidrule(r){1-1}
\cmidrule(r){2-2}
\cmidrule(r){3-6}
a [\texttt{object}] [\texttt{part}] &  19.1 & 50.7 & 18.7 & 20.7 & 10.4 \\

[\texttt{part}] of a [\texttt{object}] & 18.4 & 48.8 & 17.6 & 21.3 & 9.2   \\ 
\hline

\multirow{2}{*}{PartImageNet} & 
All & 
\multicolumn{4}{c}{quadruped} \\
& (40) & head & body & foot & tail \\
\cmidrule(r){1-1}
\cmidrule(r){2-2}
\cmidrule(r){3-6}
a [\texttt{object}] [\texttt{part}] &  29.7 & 57.3 & 25.8 & 22.9 & 22.9\\

[\texttt{part}] of a [\texttt{object}] & 29.9 & 55.9 & 25.1 & 22.9 & 24.3 \\
\shline
\end{tabular}

%% file: tables/align.tex
\tablestyle{9.5pt}{1.2}
\begin{tabular}{lccccc}
\shline
\multirow{2}{*}{Method} & 
All & 
\multicolumn{4}{c}{quadruped} \\
& (40) & head & body & foot & tail \\
\cmidrule(r){1-1}
\cmidrule(r){2-2}
\cmidrule(r){3-6}
\gc{Baseline} & \gc{5.4} & \gc{23.6} & \gc{3.4} & \gc{0.8} & \gc{1.2} \\
\cmidrule(r){1-1}
\cmidrule(r){2-2}
\cmidrule(r){3-6}
Max-score~\cite{regionclip} & 6.0 & 29.6 & 7.1 & 1.0 & 1.7  \\
Max-size~\cite{detic} & 5.3 & 20.5 & 3.5 & 0.6 & 4.7 \\
Parsed (ours) & 7.8	& 35.0	& 15.2 & 3.5 & 8.9\\

\shline
\end{tabular}

%% file: tables/foundation.tex
\tablestyle{1.5pt}{1.2}
\begin{tabular}{l l c c}
\shline
Source & capability & part name & part location \\
\cmidrule(r){1-2}
\cmidrule(r){3-3}
\cmidrule(r){4-4}
base parts & Align image and text in part-level& \checkmark & \checkmark\\
& of base objects \\
novel objects & Align image and text in object-level & \checkmark \\
&  and image-level of novel objects\\
CLIP~\cite{clip} & Anchor the part name in language & \checkmark \\
& feature space \\
DINO~\cite{dinoself}& Parse the novel object into its part &  & \checkmark \\
\shline
\end{tabular}

%% file: tables_app/stat.tex
\tablestyle{6.0pt}{1.3}
\begin{tabular}{l l r r r}
\toprule
Dataset & Type & Category & Train & Val\\
\midrule
PASCAL VOC 2007~\cite{voc}& object & 20 & 2.5k & 2.5k\\
COCO 2017~\cite{coco} & object & 80 & 118k & 5.0k \\
LVIS v1~\cite{lvis} & object & 1203 & 100k & mini 5k\\
PASCAL Part~\cite{pascalpart} & part & 93 & 4366 & 4465 \\
PartImageNet~\cite{partimagenet} & part &  40 & 16k & 2.9k\\
PACO-LVIS~\cite{paco} & part & 456 & 45k & 2.4k \\
ImageNet-1k~\cite{imagenet} & image & 1k & 1M & -\\
ImageNet-super11 & image & 11 & 16k & -\\
ImageNet-super20 & image & 20 & 49k & -\\
\bottomrule
\end{tabular}

%% file: tables_app/pascalpart_list.tex
\tablestyle{2.0pt}{1.2}
\begin{tabular}{c l l}
\toprule
Id & Name &  Part Taxonomy\\
\midrule
1 & aeroplane & body, wing, tail, wheel \\
2 & bicycle & wheel, handlebar, saddle \\
3 & bird & beak, head, eye, foot, leg, wing, neck, tail, torso \\
4 & \gc{boat} & - \\
5 & bottle & body, cap \\
6 & bus & license plate, door, headlight, mirror, window, wheel\\
7 & car & license plate, door, headlight, mirror, window, wheel\\
8 & cat & head, leg, paw, ear, eye, neck, nose, tail, torso \\
9 & \gc{chair} & - \\
10 & cow & head, leg, ear, eye, horn, muzzle, neck, tail, torso\\
11 & \gc{diningtable} & -\\
12 & dog & head, leg, paw, ear, eye, muzzle, neck, nose, tail, torso \\
13 & horse & head, leg, ear, eye, muzzle, neck, tail, torso \\
14 & motorbike & wheel, handlebar, headlight, saddle \\ 
15 & person & hair, head, ear, eye, nose, neck, mouth, arm, hand, \\
& & leg, foot, torso \\
16 & pottedplant & plant, pot\\
17 & sheep & head, leg, ear, eye, horn, muzzle, neck, tail, torso \\
18 & \gc{sofa} & -\\
19 & \gc{train} & -\\
20 & \gc{tvmonitor} & -\\
\bottomrule
\end{tabular}

%% file: tables_app/partimagenet_list.tex
\tablestyle{16.0pt}{1.2}
\begin{tabular}{c l l}
\toprule
Id & Name &  Part Taxonomy\\
\midrule
1 & Quadruped & head, body, foot, tail \\
2 & Biped & head, body, hand, foot, tail \\
3 & Fish & head, body, fin, tail\\
4 & Bird & head, body, wing, foot, tail \\
5 & Snake & head, body \\
6 & Reptile & head, body, foot, tail\\
7 & Car & body, tier, side mirror \\
8 & Bicycle & head, body, seat, tier \\
9 & Boat & body, sail \\
10 & Aeroplane & head, body, wing, engine, tail \\
11 & Bottle & body, mouth\\

\bottomrule
\end{tabular}

%% file: tables_app/partimagenet_quadruped.tex
\tablestyle{4.0pt}{1.2}
\begin{tabular}{l}
\shline 
PartImageNet: 
\texttt{quadruped} \\
\hline
Walker hound, redbone, Saluki, cairn, Boston bull, Tibetan terrier, \\
soft-coated wheaten terrier, vizsla, Brittany spaniel, English springer,\\
Irish water spaniel, Eskimo dog, chow, timber wolf, Egyptian cat,\\
leopard, tiger, cheetah, brown bear, American black bear, fox squirrel, \\
warthog, ox, water buffalo, ram, bighorn, hartebeest, impala, gazelle,\\
Arabian camel, weasel, polecat, otter, giant panda\\
\shline
\end{tabular}

%% file: tables_app/ovod_r50.tex
\tablestyle{6.5pt}{1.2}
\begin{tabular}{lcccccccccccc}
\shline
\multirow{2}{*}{Training data} &  
\multicolumn{2}{c}{VOC} & 
\multicolumn{2}{c}{COCO} & 
\multicolumn{2}{c}{LVIS} & 
\multicolumn{2}{c}{PartImageNet} &
\multicolumn{2}{c}{Pascal Part} &
\multicolumn{2}{c}{PACO}
\\
& AP & AP$_{50}$ & 
 AP & AP$_{50}$ &
 AP & AP$_{r}$ & 
 AP & AP$_{50}$ & 
 AP & AP$_{50}$ & 
 AP & AP$_{50}$ 
\\
\cmidrule(r){1-1}
\cmidrule(r){2-3}
\cmidrule(r){4-5}
\cmidrule(r){6-7}
\cmidrule(r){8-9}
\cmidrule(r){10-11}
\cmidrule(r){12-13}
LVIS, PACO  & 44.5 & 70.3 & 29.0 & 48.1 & 27.3 & 19.0 & 5.4 & 11.3 & 4.9 & 11.3 & 9.6 & 19.5\\
LVIS, PACO, Pascal Part & 42.8 & 70.8 & 28.6 & 48.0 & 26.8 & 20.4 & 7.8 & 15.3 & 21.6 & 46.3 & 9.3 & 18.9 \\
LVIS, PACO, Pascal Part, PartImageNet & 40.6 & 69.3 & 28.4 & 47.8 & 26.4 & 16.0 & 29.1 & 52.0 & 22.6 & 47.8 & 9.3 & 18.9\\
\cmidrule(r){1-1}
\cmidrule(r){2-3}
\cmidrule(r){4-5}
\cmidrule(r){6-7}
\cmidrule(r){8-9}
\cmidrule(r){10-11}
\cmidrule(r){12-13}
\gc{Dataset-specific oracle} &\gc{35.9} & \gc{69.7} & \gc{38.0} & \gc{60.8} & \gc{28.1} & \gc{20.8} & \gc{29.7} & \gc{54.1} & \gc{19.4} & \gc{42.3} & \gc{10.6} & \gc{21.7} \\
\shline
\end{tabular}

%% file: tables_app/ovod_swinbase.tex
\tablestyle{6.5pt}{1.2}
\begin{tabular}{lcccccccccccc}
\shline
\multirow{2}{*}{Training data} &  
\multicolumn{2}{c}{VOC} & 
\multicolumn{2}{c}{COCO} & 
\multicolumn{2}{c}{LVIS} & 
\multicolumn{2}{c}{PartImageNet} &
\multicolumn{2}{c}{Pascal Part} &
\multicolumn{2}{c}{PACO}
\\
& AP & AP$_{50}$ & 
 AP & AP$_{50}$ &
 AP & AP$_{r}$ & 
 AP & AP$_{50}$ & 
 AP & AP$_{50}$ & 
 AP & AP$_{50}$ 
\\
\cmidrule(r){1-1}
\cmidrule(r){2-3}
\cmidrule(r){4-5}
\cmidrule(r){6-7}
\cmidrule(r){8-9}
\cmidrule(r){10-11}
\cmidrule(r){12-13}
LVIS, PACO  & 55.2 & 72.2 & 41.0 & 58.4 & 41.3 & 32.8 & 6.9 & 13.7 & 5.6 & 12.5 & 15.9 & 31.9 \\
LVIS, PACO, Pascal Part & 52.6 & 72.4 & 40.4 & 57.9 & 39.9 & 29.8 & 11.8 & 21.8 & 30.5 & 59.3 & 15.4 & 30.2 \\
LVIS, PACO, Pascal Part, PartImageNet & 50.3 & 71.6 & 40.3 & 57.8 & 39.6 & 30.3 & 40.0 & 64.8 & 31.2 & 60.5 & 15.4 & 30.3 \\
\cmidrule(r){1-1}
\cmidrule(r){2-3}
\cmidrule(r){4-5}
\cmidrule(r){6-7}
\cmidrule(r){8-9}
\cmidrule(r){10-11}
\cmidrule(r){12-13}
\gc{Dataset-specific oracle} & 
\gc{59.0} & \gc{82.0} & \gc{52.5} & \gc{72.0} & \gc{43.1} & \gc{38.7} & \gc{41.7} & \gc{68.7} & \gc{27.4} & \gc{56.1} & \gc{15.2} & \gc{29.4} \\
\shline
\end{tabular}

%% file: tables_app/pascalpart_list_ori.tex
\tablestyle{5.0pt}{1.2}
\begin{tabular}{c l l}
\toprule
Id & Name &  Part Taxonomy\\
\midrule
1 & aeroplane & body, engine, left wing, right wing, stern, tail, wheel \\
2 & bicycle & back wheel, chain wheel, front wheel, handlebar, headlight, saddle \\
3 & bird & beak, head, left eye, left foot, left, leg, left wing, neck, right eye, right foot, right leg, right wing, tail, torso \\
4 & boat & - \\
5 & bottle & body, cap \\
6 & bus & back license plate, back side, door, front license plate, front side, headlight, left mirror, left side, right mirror, \\
& &
right side, roof side, wheel, window \\
7 & car & back license plate, back side, door, front license plate, front side, headlight, left mirror, left side, right mirror, \\
& &
right side, roof side, wheel, window \\
8 & cat & head, left back leg, left back paw, left ear, left eye, left front leg, left front paw, neck, nose, right back leg, \\
& &
right back paw, right ear, right eye, right front leg, right front paw, tail, torso \\
9 & chair & - \\
10 & cow & head, left back lower leg, left back upper leg, left ear, left eye, left front lower leg, left front upper leg,  \\
& & 
left horn, muzzle, neck, right back lower leg, right back upper leg, right ear, right eye, right front lower leg, \\
& & 
right front upper leg, right horn, tail, torso\\
11 & diningtable & -\\
12 & dog & head,
left back leg,
left back paw,
left ear,
left eye,
left front leg,
left front paw,
muzzle,
neck,
nose,
right back leg, \\ 
& &
right back paw,
right ear,
right eye,
right front leg,
right front paw,
tail,
torso
\\
13 & horse &
head,
left back hoof,
left back lower leg,
left back upper leg,
left ear,
left eye,
left front hoof,
left front lower leg, \\ 
& &
left front upper leg,
muzzle,
neck,
right back hoof,
right back lower leg,
right back upper leg,
right ear,
right eye,\\
& & 
right front hoof,
right front lower leg,
right front upper leg,
tail,
torso
\\
14 & motorbike &
back wheel,
front wheel,
handlebar,
headlight,
saddle
\\ 
15 & person & hair,
head,
left ear,
left eye,
left eyebrow,
left foot,
left hand,
left lower arm,
left lower leg,
left upper arm, \\ 
& &
left upper leg, mouth, neck, nose, right ear, right eye, right eyebrow, right foot, right hand, right lower arm, \\
& & 
right lower leg, right upper arm, right upper leg, torso \\
16 & pottedplant & plant, pot\\
17 & sheep & head, left back lower leg, left back upper leg, left ear, left eye, left front lower leg, left front upper leg, \\ 
& & 
left horn, muzzle, neck, right back lower leg, right back upper leg, right ear, right eye, right front lower leg, \\
& & 
right front upper leg, right horn, tail, torso \\
18 & sofa & -\\
19 & train & coach back side, coach front side, coach left side, coach right side, coach roof side, coach, head, head back side, \\ 
& &
head front side, head left side, head right side, head roof side, headlight\\
20 & tvmonitor & screen\\
\bottomrule
\end{tabular}

%% file: tables_app/paco_list.tex
\tablestyle{2.0pt}{0.8}
\begin{tabular}{r l l}
\toprule
Id  &  Name  &  Part Taxonomy\\
\midrule
1 & ball & - \\
2  &  basket & bottom, handle, inner\_side, cover, side, rim, base\\
3  &  belt & buckle, end\_tip, strap, frame, bar, prong, loop, hole\\
4  &  bench & stretcher, seat, back, table\_top, leg, arm\\
5  &  bicycle & stem, fork, top\_tube, wheel, basket, seat\_stay, saddle, handlebar, pedal, gear, head\_tube, down\_tube, seat\_tube\\
6  &  blender & cable, handle, cover, spout, vapour\_cover, base, inner\_body, seal\_ring, cup, switch, food\_cup\\
7  &  book & page, cover\\
8  &  bottle & neck, label, shoulder, body, cap, bottom, inner\_body, closure, heel, top, handle, ring, sipper, capsule, spout, base, punt\\
9  &  bowl & inner\_body, bottom, body, rim, base\\
10  &  box & bottom, lid, inner\_side, side\\
11  &  broom & lower\_bristles, handle, brush\_cap, ring, shaft, brush\\
12  & bucket & handle, cover, body, base, inner\_body, bottom, loop, rim\\
13  & calculator & key, body\\
14  & can & pull\_tab, body, base, inner\_body, bottom, lid, text, rim\\
15  & car\_(automobile) & headlight, turnsignal, tank, windshield, mirror, sign, wiper, fender, trunk, windowpane, seat, logo, grille, antenna, hood, \\
 &  &  splashboard, bumper, rim, handle, runningboard, window, roof, wheel, taillight, steeringwheel\\
16  & carton & inner\_side, tapering\_top, cap, bottom, lid, text, side, top\\
17  & cellular\_telephone & button, screen, bezel, back\_cover\\
18  & chair & stretcher, swivel, apron, wheel, leg, base, spindle, seat, back, rail, stile, skirt, arm\\
19  & clock & cable, decoration, hand, pediment, finial, case, base\\
20  & crate & bottom, handle, inner\_side, lid, side\\
21  & cup & inner\_body, handle, rim, base\\
22 & dog & teeth, neck, foot, head, body, nose, leg, tail, ear, eye\\
23 & drill & handle, body\\
24 & drum\_(musical\_instrument) & head, rim, cover, body, loop, lug, base\\
25 & earphone & headband, cable, ear\_pads, housing, slider\\
26 & fan & rod, canopy, motor, blade, base, string, light, bracket, fan\_box, pedestal\_column\\
27 & glass\_(drink\_container) & inner\_body, bottom, body, rim, base\\
28 & guitar & key, headstock, bridge, body, fingerboard, back, string, side, pickguard, hole\\
29 & hammer & handle, face, head, grip\\
30 & handbag & zip, inner\_body, handle, bottom, body, rim, base\\
31 & hat & logo, pom\_pom, inner\_side, strap, visor, rim\\
32 & helmet & face\_shield, logo, inner\_side, strap, visor, rim\\
33 & jar & handle, body, base, inner\_body, bottom, lid, sticker, text, rim\\
34 & kettle & cable, handle, lid, body, spout, base\\
35 & knife & handle, blade\\
36 & ladder & rail, step, top\_cap, foot\\
37 & lamp & shade\_inner\_side, cable, pipe, shade, bulb, shade\_cap, base, switch, finial\\
38 & laptop\_computer & cable, camera, base\_panel, keyboard, logo, back, screen, touchpad\\
39 & microwave\_oven & inner\_side, door\_handle, time\_display, control\_panel, turntable, dial, side, top\\
40 & mirror & frame\\
41 & mouse\_(computer\_equipment) & logo, scroll\_wheel, body, right\_button, wire, side\_button, left\_button\\
42 & mug & handle, body, base, inner\_body, bottom, text, drawing, rim\\
43 & napkin & -\\ 
44 & newspaper & text\\
45 & pan\_(for\_cooking) & bottom, handle, inner\_side, lid, side, rim, base\\
46 & pen & cap, grip, barrel, clip, tip\\
47 & pencil & body, lead, eraser, ferrule\\
48 & pillow & embroidery\\
49 & pipe & nozzle, colied\_tube, nozzle\_stem\\
50 & plastic\_bag & inner\_body, handle, text, hem, body\\
51 & plate & top, bottom, inner\_wall, body, rim, base\\
52 & pliers & jaw, handle, joint, blade\\
53 & remote\_control & logo, back, button\\
54 & scarf & fringes, body\\
55 & scissors & handle, screw, finger\_hole, blade\\
56 & screwdriver & blade, handle, tip, shank\\
57 & shoe & toe\_box, tongue, vamp, outsole, insole, backstay, lining, quarter, heel, throat, eyelet, lace, welt\\
58 & slipper\_(footwear) & toe\_box, vamp, outsole, strap, insole, lining\\
59 & soap & neck, label, shoulder, body, sipper, capsule, spout, push\_pull\_cap, cap, base, bottom, closure, punt, top\\
60 & sponge & rough\_surface\\
61 & spoon & neck, handle, bowl, tip\\
62 & stool & seat, leg, step, footrest\\
63 & sweater & shoulder, sleeve, neckband, hem, body, yoke, cuff\\
64 & table & stretcher, drawer, inner\_wall, shelf, apron, wheel, leg, top, rim\\
65 & tape\_(sticky\_cloth\_or\_paper) & roll\\
66 & telephone & button, screen, bezel, back\_cover\\
67 & television\_set & bottom, button, side, top, base\\
68 & tissue\_paper & roll\\
69 & towel & body, terry\_bar, hem, border\\
70 & trash\_can & label, body, wheel, inner\_body, bottom, lid, pedal, rim, hole\\
71 & tray & bottom, inner\_side, outer\_side, rim, base\\
72 & vase & neck, handle, foot, body, mouth\\
73 & wallet & inner\_body, flap\\
74 & watch & buckle, case, dial, hand, strap, window, lug\\
75 & wrench & handle, head\\
\bottomrule
\end{tabular}